\documentclass{article}
\usepackage{amssymb}
\usepackage{amsmath}
\usepackage{amsthm}

\newtheorem{thm2}{Theorem}

\newtheorem{deff}[thm2]{Definition}
\usepackage{a4wide}
\usepackage{multirow,graphicx,subfigure}
\usepackage{bm}
\usepackage{subfigure}

\def\R{\in \mathbb{R}}

\def\cov{\text{Cov}}

\def\lv{\left\Vert}
\def\rv{\right\Vert}

\def\dx{\frac{\partial}{\partial \bm{x}_i}}

\newcommand{\tr}{^{\top}}

\begin{document}

\title{Multi-view predictive partitioning in high dimensions}

\author{Brian McWilliams and Giovanni Montana\footnote{To whom correspondence should be addressed. Email: g.montana@imperial.ac.uk} \\Statistics Section, Department of Mathematics\\Imperial College London, UK}

\maketitle

\begin{abstract}
Many modern data mining applications are concerned with the analysis of datasets in which the observations are described by paired high-dimensional vectorial representations or ``views''. Some typical examples can be found in web mining and genomics applications. In this article we present an algorithm for data clustering with multiple views, Multi-View Predictive Partitioning (MVPP), which relies on a novel criterion of predictive similarity between data points. We assume that, within each cluster, the dependence between multivariate views can be modelled by using a two-block partial least squares (TB-PLS) regression model, which performs dimensionality reduction and is particularly suitable for high-dimensional settings. The proposed MVPP algorithm partitions the data such that the within-cluster  predictive ability between views is maximised. The proposed objective function depends on a measure of predictive influence of points under the TB-PLS model which has been derived as an extension of the PRESS statistic commonly used in ordinary least squares regression. Using simulated data, we compare the performance of MVPP to that of competing multi-view clustering methods which rely upon geometric structures of points, but ignore the predictive relationship between the two views. State-of-art results are obtained on benchmark web mining datasets. 
\end{abstract}

\section{Introduction}

In recent years, an increasing number of data mining applications have arisen which deal with the problem of finding patterns in data points for which several blocks of high-dimensional measurements, or ``views'', have been obtained. Each view generally provides a different quantitative representation of the available random samples. For instance, in genomics, one view may represent the expression levels of all genes in the genome, and the paired view may represent the genetic alternations in each gene observed on the same biological samples \cite{Witten2009a}; in web mining applications, web pages may be represented by the text content of the page and the hyperlinks pointing to it \cite{Bickel2004}. 
The fast-developing area of multi-view learning \cite{Caruana1997, Cesa-Bianchi2010} is concerned with how to make combined use of the information provided by the available views in order to perform specific data mining tasks, such as clustering and predictive modelling.  

A recurrent objective is that of detecting naturally occurring data clusters. 
When multiple views are available, there may be reasons to believe that the observations should cluster in the same way under each of the available views, hence the process of inferring the true clustering can be improved by making joint use of all views. 
 A growing number of multi-view clustering algorithms have been proposed, and two main approaches seem to have emerged in the literature: {\it late} and {\it early} fusion methods. The late fusion methods first recover the clusters independently from each view (e.g. by using the K-means algorithm), and then attempt to infer a ``consensus'' clustering by combining the partitioning obtained within each view such that some measure of disagreement between the individual partitionings is minimised \cite{Li,Lange2006,Long2008,Bruno2009}. 

On the other hand, early fusion methods start by learning any common patterns that may be shared by the views and that could yield a joint clustering \cite{Bickel2004,Tzortzis2009,Chaudhuri2009,Sa2010}. A common assumption is that the data under each view are generated from a mixture of distributions where the mixing proportions which determine the cluster assignments are unknown but shared between views. 
 Several methods rely on a two-step approach where the clusters are ultimately defined in terms of geometric separation between points lying in low-dimensional projections: first, a joint dimensionality reduction step is performed using both views, generally using canonical correlations analysis (CCA) which recovers latent factors explaining the correlation between the views; second, K-means is used to detect the clusters among data points in the lower dimensional space found in the first stage. Using CCA to initially perform dimensionality reduction has been shown to increase the separation between the clusters \cite{Chaudhuri2009}, and some non-linear extensions have also been explored \cite{Sa2010}. 

These multi-view clustering methods have become particularly popular for a broad class of problems in web mining where data of many different types may co-occur on the same page, for example text, hyperlinks, images and video. A common application is that of clustering web pages \cite{Bickel2004}. Pages can be clustered together based on the similarity of both their text content and link structure, and the clusters identify broad subject categories. Another problem in the web mining domain involves clustering and annotating images represented as a vector of pixel intensities as well as bag of words containing the corresponding textual annotation \cite{Bruno2009}. The ability to accurately cluster these data have implications in web search and advertising.

A different multi-view learning scenario arises when one representation of the data is treated as a high-dimensional predictor or ``explanatory'' vector, and the paired view represents a high-dimensional ``response'' vector. The task then consists in fitting a regression model such that, when a new observation has been observed under the explanatory view, the corresponding representation of that observation in the response view can be optimally predicted. 


In settings when both the explanatory and response views are high-dimensional, two-block partial least squares (TB-PLS) regression has proved to be a particularly useful method for modelling a linear predictive relationship between two high-dimensional views \cite{Wegelin2000,Rosipal2006}. TB-PLS performs dimensionality reduction in both predictor and response views simultaneously by assuming that each multivariate representation of the data can be factorized in a set of mutually orthogonal latent factors that maximise the covariance between the views. This regression model overcomes problems relating to multicollinearity by estimating least squares regression coefficients using the lower dimensional projections. 
Among other applications, the model has been successfully used in genomics, where regions of DNA that are highly predictive of gene expression need to be detected \cite{LeCao2009}, and in computational finance, where the returns of several indices have been predicted using a large basket of assets \cite{McWilliams2010}.

In this article we also assume that two views are available, and set out to discover any potential partitions of the data by using them jointly. 
Whereas other multi-view clustering approaches rely on geometrical structures, we assume that any two points should be assigned to the same cluster if they both appear to be modelled equally well by the same regression model. In this respect, multi-view clustering is framed as a problem of learning the unknown number of multi-response regression models in high-dimensions. This is accomplished by first introducing a novel criterion for quantifying the \emph{predictive influence} of an observation under a TB-PLS regression model, whereby a data point is deemed {\it unusual} for the model if it has high predictive influence. The rationale is that, under a given cluster-specific regression model, any unusual observation should be removed from that cluster and allocated to a different one. 

The article is organised as follows. In Section \ref{pls} we review the two-block PLS regression model, and describe the problem of modelling heterogeneous data. In Section \ref{sec_pred} we introduce a measure of predictive influence for TB-PLS regression, and address the predictive partitioning problem; an objective function is first proposed, and an iterative {\it multi-view predictive partitioning} (MVPP) algorithm is presented. To be best of our knowledge, no other multi-view {\it predictive} clustering algorithm has been proposed in the literature.  In Section \ref{sec_simsettings} we describe a number of Monte Carlo simulation settings that will be used   to illustrate the performance of the proposed methods under different scenarios. The results, as well as comparisons to alternative multi-view clustering algorithms, are then presented in Section \ref{sec_results}. The applications to real web pages and academic paper clustering in Section \ref{sec_results2} demonstrate the performance of the algorithm on real data. Concluding remarks are found in Section \ref{sec_conc}.%

\section{High-dimensional multi-response regression} \label{pls}

\subsection{Two block partial least squares regression}
\label{sec_pls}

Suppose we have observed a random sample of $n$ independent and identically distributed data points, $\{\bm{x}_i, \bm{y}_i\}$, for $i=1,\ldots,n$, where each $\bm{x}_i \R^{1\times p}$ is the ``explanatory view'' and each $\bm{y}_i \R^{1\times q}$ is the ``response view'' observed on the $i^{\text{th}}$ sample. The dimensions of both views, $p$ and $q$, are allowed to be very large. The $n$ observations can then be arranged in two paired matrices, one containing all the explanatory variables observed in the samples, $\bm{X}\R^{n\times p}$, and one containing the corresponding response variables, $\bm{Y}\R^{n\times q}$. The variables in both views are centred and scaled so as to have zero mean and unit variance.

The TB-PLS regression model assumes that the predictor and response views are noisy realisations of linear combinations of hidden variables, or {\it latent factors}.
The specific form of the TB-PLS model is given in the following definition.
\begin{deff}
The TB-PLS model assumes the existence of $R$ pairs of orthogonal latent factors, $\bm{t}^{(r)}$ and  $\bm{s}^{(r)} \R^{n\times 1}$, for $r=1,\ldots,R$ such that
\begin{align} \label{pls_model}
\bm{X} = \sum_{r=1}^{R} \bm{t}^{(r)} {\bm{p}^{(r)}}\tr + \bm{E}_x,  ~~~ 
\bm{Y} = \sum_{r=1}^{R} \bm{s}^{(r)} {\bm{q}^{(r)}}\tr + \bm{E}_y, 
\end{align} 
where $\bm{E}_x \R^{n\times p}$ and $\bm{E}_y \R^{n \times q}$ are matrices of residuals. 
For each $r$, the latent factors  are 
$\bm{t}^{(r)}=\bm{X}\bm{u}^{(r)}$ and $\bm{s}^{(r)}=\bm{Y}\bm{v}^{(r)}$ where $\bm{u}^{(r)}\R^{p\times 1}$ and $\bm{v}^{(r)}\R^{q\times 1}$ are weight vectors of unit length. The vectors $\bm{p}^{(r)} \R^{p\times 1}$ and $\bm{q}^{(r)} \R^{q\times 1}$ are the {\it factor loadings}.
\label{def_tbpls}
\end{deff}
For any given $r$, each pair of latent factors $\{\bm{t}^{(r)}, \bm{s}^{(r)}\}$ provides a one-dimensional representation of both views and is obtained by identifying the directions on which the projected views have maximal covariance. Therefore, the paired latent factors satisfy the property that 
	\begin{equation} \label{eq_pls_of}
			 \cov (\bm{t}^{(r)},\bm{s}^{(r)}) = \max_{\bm{u}^{(r)},\bm{v}^{(r)}} \cov (\bm{Xu}^{(r)},\bm{Yv}^{(r)})^2,
	\end{equation} 
under the constraints that $\lv \bm{u}^{(r)} \rv = \lv \bm{v}^{(r)} \rv =1$ for all $r=1,\ldots,R$. 
For $r=1$ this optimisation problem is equivalent to
\begin{equation} \label{eq_plssvd}
\lambda^{(1)} = {\bm{t}^{(1)}}\tr\bm{s}^{(1)} = \max_{\bm{v}^{(1)}, \bm{u}^{(1)}} {\bm{u}^{(1)}}\tr \bm{X}\tr \bm{Y} \bm{v}^{(1)}
\end{equation}
under the same constraints posed on the weights. Here, $\lambda^{(1)}$ is the largest singular value of $\bm{X}\tr\bm{Y}$ and the weights are the corresponding left and right singular vectors. The $R$ weight vectors that satisfy Eq. \eqref{eq_pls_of} can then be found by computing the singular value decomposition (SVD) of $\bm{X}\tr\bm{Y}$, that is $\bm{X}\tr\bm{Y}=\bm{U} \Lambda \bm{V} \tr$, where $\bm{U}=[ \bm{u}^{(1)},...,\bm{u}^{(p)}] \R^{p \times p}$ and $\bm{V}=[ \bm{v}^{(1)},...,\bm{v}^{(q)}] \R^{q \times q}$ are orthonormal matrices whose columns are the left and right singular vectors of $\bm{X}\tr\bm{Y}$, respectively. $\Lambda \R^{p \times q}$ is a diagonal matrix whose entries are the ordered singular values of $\bm{X}\tr\bm{Y}$. Therefore $\bm{u}^{(r)}$ and $\bm{v}^{(r)}$ are taken to be the $r^{\text{th}}$ left and right singular vectors of $\bm{X}\tr\bm{Y}$, respectively.  

The predictive relationship between the two views is driven by a linear regression model involving the $R$ pairs of latent factors. For each $r$, the response latent variable depends on the explanatory latent variable, as follows
\begin{equation} \label{eq_lf_residual}
\bm{s}^{(r)}=\bm{t}^{(r)} g^{(r)} + \bm{h}^{(r)}  
\end{equation}
where each $g^{(r)}$ is a scalar regression coefficient which describes the projection of the latent factor relating to the response onto the latent factor relating to the predictors, and each $\bm{h}^{(r)}\R^{n\times 1}$ is the vector of residual errors. Since the latent factors are assumed to have zero mean, there is no intercept term. Using the inner regression models \eqref{eq_lf_residual}, the TB-PLS model \eqref{pls_model} can now be re-written in the more familiar form
\begin{align} \label{eq_pls_beta}
			\bm{Y} &
= \bm{X} \sum_{r=1}^{R}\bm{u}^{(r)} g^{(r)} {\bm{q}^{(r)}}\tr + \bm{E} 
					 = \bm{X} \bm{\beta} + \bm{E}, 
\end{align}
where the regression coefficients have been defined as
\begin{equation} \label{eq:coeff}
\bm{\beta} = \sum_{r=1}^{R}\bm{u}^{(r)} g^{(r)} {\bm{q}^{(r)}}\tr
\end{equation}
and depends on the parameter sets $\bm{\Theta}^{(r)}=\{\bm{u}^{(r)}, \bm{v}^{(r)} , g^{(t)}, \bm{q}^{(r)}\}$, with $r=1, \ldots, R$.  Each one of the $R$ factor loadings $\bm{q}^{(r)}$ are obtained by performing univariate regressions,
\begin{equation}
\bm{q}^{(r)} = \frac{ \bm{Y}\tr {\bm{s}^{(r)}}}{{\bm{s}^{(r)}}\tr \bm{s}^{(r)}} .
\label{eq_q}
\end{equation}
and each of the $R$ regression coefficients $g^{(r)}$, from the inner model of Eq. \eqref{eq_lf_residual}, is estimated by least squares regression of $\bm{t}^{(r)}$ on $\bm{s}^{(r)}$, so that
\begin{equation}
g^{(r)} = {\left({\bm{t}^{(r)}}\tr \bm{t}^{(r)}\right)}^{-1}{\bm{t}^{(r)}}\tr \bm{s}^{(r)}. \label{eq_g}
\end{equation}

In high-dimensional settings, such as the one we consider, it is generally appropriate to assume the data has spherical covariance within each view \cite{Chaudhuri2009}, and so $\bm{X}\tr \bm{X} = I_p$ and $\bm{Y}\tr \bm{Y} = I_q$. Although  clearly incorrect in many real-world applications, this assumption has been shown to provide better results, particularly in classification problems, than attempting to estimate the true covariance matrices especially when $p,q>>n$ \cite{Bickel2004b}. This can be seen as an extreme form of regularisation which introduces a large bias to reduce the variance in the estimated parameters and has been widely used in applications involving genomic data  \cite{Tibshirani2003,Dudoit2000,Parkhomenko2009,Witten2009a}.

\subsection{Modelling heterogeneous data} \label{sec_heterogeneous}

The TB-PLS regression model rests on the assumption that the $n$ independent samples are representative of a single, homogeneous population. 
Under this assumption, the latent factors that determine the regression coefficients in Eq. \eqref{eq:coeff} can be optimally estimated using all the available data. 
However, in many applications the observations may be representative of a number of different populations, each one characterised by a different between-views covariance structure. Failing to recognise this would lead to a biased estimation of the latent factors, which would in turn provide a sub-optimal predictive model. 

We are interested in situations in which the observations have been sampled from $K$ different sub-populations, where the exact value of $K$ may be unknown. 
It can be noted that in general the optimal dimension $R_k$ is not necessarily the same across clusters. The problem involves simultaneously recovering the cluster assignments and their parameter sets, as well as learning the optimal $K$. Learning the optimal dimensionality in each cluster is a much harder problem which we address later.

\begin{figure}[htp] 
\subfigure[Simulated data.]{
\centerline{\includegraphics[width=1\columnwidth]{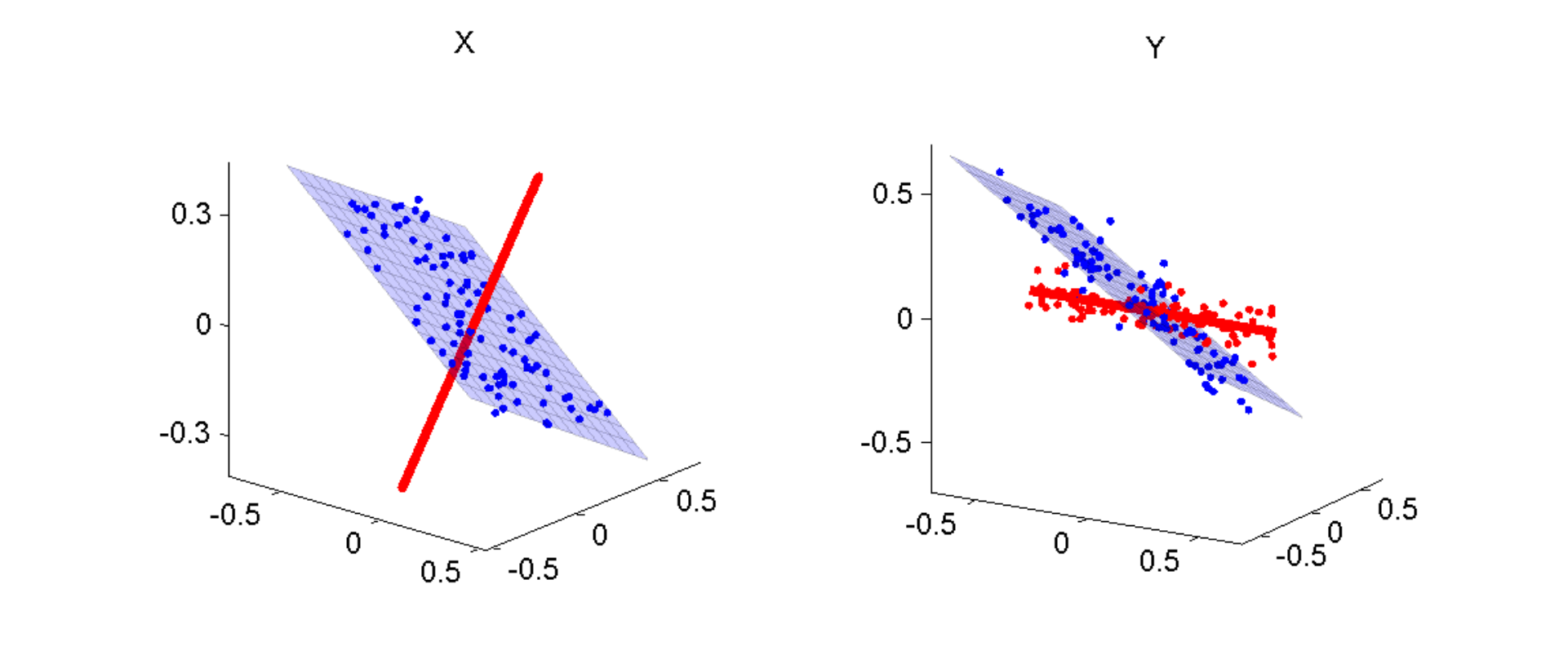}}
\label{fig:subfig1}}
\subfigure[Global TB-PLS.]{
\centerline{\includegraphics[width=1\columnwidth]{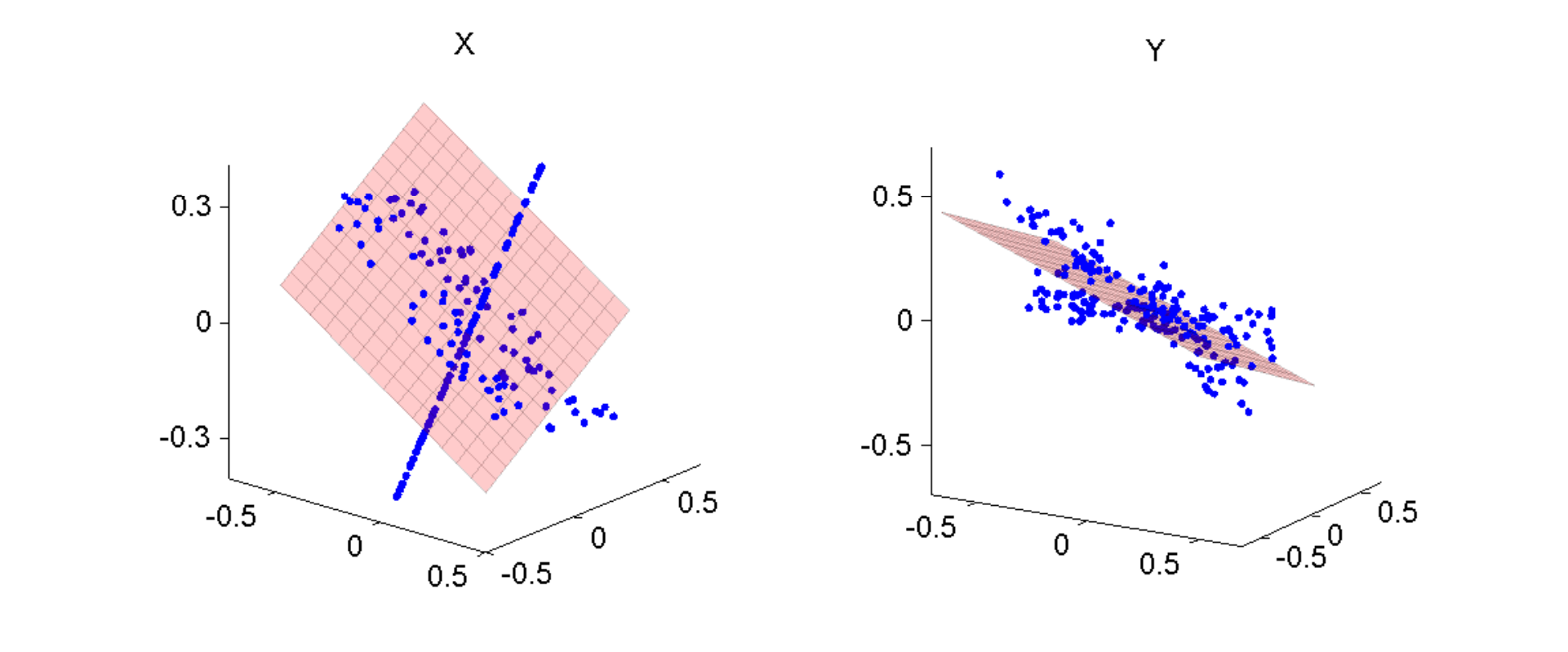}}
\label{fig:subfig2}}
\caption{Figure \ref{fig:subfig1} shows the two clusters in the $\bm{X}$ view consist of points sampled uniformly on a line and a plane embedded in three dimensions. The clusters in the $\bm{Y}$ view  are noisy linear combinations of the corresponding clusters in the $\bm{X}$ view so that there is a predictive relationship between the views. Figure \ref{fig:subfig2} shows the result of fitting a global TB-PLS model to the data. It can be seen that the resulting subspace in the $\bm{X}$ view lies between the clusters and as a result few of the observations in the response lie on the estimated subspace.}
\label{fig_subspace}
\end{figure}

A simple illustrative example is given in Figure \ref{fig:subfig1} where $K=2$, $R_1=2$, $R_2=1$ and $p=3$. Here, under the $\bm{X}$ view, the points are uniformly distributed along either one of two lower dimensional subspaces, a line and a plane, both embedded in the three-dimensional space. To generate data points under the $\bm{Y}$ view that can be linearly predicted using the explanatory view, we take a linear combination of variables in the explanatory view and add some Gaussian noise. Clearly, fitting a global TB-PLS model would be inappropriate here, as shown in Figure \ref{fig:subfig2}, which shows that the estimated subspaces differ from the true ones, so the predictive ability of the model is sub-optimal. 
We will revisit this example in Section \ref{sec_results} and show that our multi-view clustering algorithm recovers the true sub-spaces, as in Figure \ref{fig_subspace2}.

\marginpar{[{\bf FIGURE \ref{fig_subspace} AROUND HERE}]}
 



\section{Predictive partitioning} \label{sec_pred}

\subsection{A PRESS-based predictive influence measure}

The issue of detecting influential observations has been extensively studied in the context of OLS regression \cite{Belsley1980,Meloun2001}. A common approach is based on examining the elements of the $(n \times n)$ ``hat matrix'', $\bm{G}=\bm{X}(\bm{X}\tr \bm{X})^{-1}\bm{X}\tr $.
The term $G_{ii}$ is known as the leverage of the $i^{th}$ point, and determines the contribution of the $i^{th}$ point in estimating its associated response. The partial leverage, $G_{ij}$ is a related quantity which gives the contribution of the $j^{th}$ point for estimating the response associated to the $i^{th}$ point. These quantities are used to detect influential observations which have a larger contribution to the estimated responses relative to other points. However, $\bm{G}$ does not take into account any information from $\bm{Y}$ and so these leverage terms alone are not always sufficient to determine which observations are influential \cite{Belsley1980}. 

After fitting the regression model, a seemingly obvious way to identify influential observations might be to examine the residual error. However, it has been observed that points which exert a large leverage on the regression may obtain relatively smaller residual errors compared to other points as a result of over-fitting \cite{Pregibon1981}.

A more effective approach to assessing the influence of a particular observation considers the effects of its removal from the regression model.  This involves estimating the regression parameter $n$ times, leaving out each observation in turn, and then evaluating the prediction error on the unused observation. If we let $\bm{\beta}_{-i}$ be the OLS regression coefficient estimated by using all but the $i^{th}$ observation, the corresponding leave-one-out (LOO) error is $\bm{e}_{-i} = \bm{y}_{i}-\bm{x}_i \bm{\beta}_{-i}$. An observation can then be labelled as influential if its LOO error is particularly large. The choice of threshold for identifying an observation as influential is an open question and many strategies have been suggested in the literature \cite{Belsley1980}. 

The approach above is related to the leave-one-out cross validation error (LOOCV)  which is often used to quantify the predictive performance of a regression model \cite{Stone1974a}, and is defined as the mean of the individual prediction errors, 
\begin{equation}
J = \frac{1}{n} \sum_{i=1}^{n}\lv \bm{e}_{-i} \rv ^{2}.\label{eq_press}
\end{equation}
For OLS, it is well known that each prediction error featuring in Eq. \eqref{eq_press} can be computed without the need to remove an observation and re-fit the regression model. This can be accomplished through a closed-form expression known as the PRESS \cite{Belsley1980}, which gives 
\begin{align}
\bm{e}_{-i} = \frac{\bm{e}_i}{1-{G}_{ii}}.
\label{eq_olspress}
\end{align}

In this form, the $i^{th}$ leave-one-out residual can be seen as the $i^{th}$ residual, $\bm{e}_i$ scaled by one minus its leverage, ${G}_{ii}$. Since the PRESS only depends on quantities estimated using least squares it has a computational cost in the order of a single least squares fit and, as such, is extremely efficient to compute. 

In previous studies, the PRESS has also been used for identifying influential observations in the context of PLS regression with univariate responses \cite{Wold1984,Martens2000}. However, in practice its computation requires the regression model to be fit $n$ times, each time using $n-1$ data points. A similar strategy for the evaluation of the PRESS in an TB-PLS model, when the response is multivariate, would require $n$ SVD computations, each one having a computational cost of $O\left(p^2q + q^2p\right)$ \cite{Golub1996}. This approach is particularly expensive when the dimension of the data in either view is large, as in our settings.  

Recently, we proposed a closed-form expression for computing the PRESS statistic under a TB-PLS model which reduces the computational cost of explicitly evaluating the leave-one-out errors \cite{McWilliams2010a}. We overcome the need to recompute the SVD $n$ times by approximating the leave-one-out estimates of the singular vectors $\{ \bm{u}_{-i}, \bm{v}_{-i} \}$ with $\{ \bm{u}, \bm{v} \}$.
\begin{deff}
A closed-form approximation for the PRESS in Eq. \eqref{eq_press} in a TB-PLS model is
\begin{align}
J \approx \frac{1}{n} \sum_{i=1}^n \lv \frac{ \bm{e}_i -t_i^2 \bm{E}_{y,i} - \bm{b}_i }{(1-t_i^2)(1-s_i^2)} \rv^2 , \label{eq_press_refac}
\end{align}
where $\bm{e}_i = \bm{y}_i-\bm{x}_i\bm{\beta}$ is the TB-PLS residual error, and $\bm{b}= h_i s_i \bm{y}_i$, with $h_i= s_i -g t_i$ being the $i^{th}$ residual error for the inner regression model of Eq. \eqref{eq_lf_residual} and  $\bm{E}_{y,i} \R^{1\times q}$ being the $i^{th}$ residual in the TB-PLS model in Eq. \eqref{pls_model}. 
\label{def_press}
\end{deff}

The derivation of Eq. \eqref{eq_press_refac} is provided in Appendix A. The error introduced by approximating the leave-one-out estimates of the singular vectors is of order $O\left(\sqrt{\frac{\log (n)}{n}}\right)$. The denominator of  Eq. \eqref{eq_press_refac} is a scaling term related to the contribution of each data point to the latent factors, $\bm{t}$ and $\bm{s}$. In this form, it can be seen that the TB-PLS PRESS has similarities with the PRESS for OLS regression in Eq. \eqref{eq_olspress} where these scaling terms are related to the leverage each point exerts on the regression \cite{Belsley1980}. 
 
Using Eq. \eqref{eq_press_refac}, we now consider how to measure the influence each point exerts on the TB-PLS model. Since we are interested in the predictive performance of the TB-PLS model, we aim to identify influential points as those observations having the greatest effect on the prediction error.  In order to quantify this effect, we define the \emph{predictive influence} of an observation $\{\bm{x}_i, \bm{y}_i \}$ as the rate of change of the PRESS at that point whilst all other observations remain constant. 
\begin{deff}
The predictive influence of a data point $\{ \bm{x}_i, \bm{y}_i\}$,  which we denote as $\bm{\pi}_{\bm{x},\bm{y}} (\bm{x}_i, \bm{y}_i)\R^{(p+q)\times 1}$, is the total derivative of the PRESS with respect to the $p$ variables in $\bm{x}_i$ and the $q$ variables in $\bm{y}_i$,
\begin{align}
\bm{\pi}_{\bm{x},\bm{y}} (\bm{x}_i, \bm{y}_i)  = \left[ \frac{\partial J}{\partial \bm{x}_{i,1}}, \ldots , \frac{\partial J}{\partial \bm{x}_{i,p}}, ~ \frac{\partial J}{\partial \bm{y}_{i,1}}, \ldots , \frac{\partial J}{\partial \bm{y}_{i,q}}  \right]\tr .
\end{align}
\end{deff}

The closed-form expression for the computation of this quantity is reported in Appendix \ref{sec_appendixB}. The predictive influence offers a way of measuring how much the prediction error would increase in response to an incremental change in the observation $\{\bm{x}_i, \bm{y}_i \}$ or alternatively, the sensitivity of the prediction error with respect to that observation. The rate of change of the PRESS at this point is given by the magnitude of the predictive influence vector, $\lv \bm{\pi}_{\bm{x},\bm{y}} (\bm{x}_i, \bm{y}_i)\rv^2$. If this quantity is large, this implies a small change in the observation will result in a large change in the prediction error relative to other points. In this case, removing such a point from the model would cause a large improvement in the prediction error. We can then identify influential observations as those for which the increase in the PRESS is large, relative to other observations. 

In the remainder of this Section we develop further the idea of using the predictive influence measure for multi-view clustering. 

\subsection{The MVPP clustering algorithm} \label{sec_pred_dist}

Initially we assume that the number of clusters, $K$, is known. 
As mentioned in Section \ref{sec_heterogeneous}, we want to allocate each observation $\{ \bm{x}_i, \bm{y}_i\}$, $i=1,\ldots,n$ into one of $K$ non-overlapping clusters $\{\mathcal{C}_1,\ldots,\mathcal{C}_K \}$ such that each cluster contains exactly $n_k$ observations, with $\sum_{k=1}^K n_k=n$, and these points are as similar as possible in a predictive sense. Accordingly, we first define a suitable objective function to be minimised.
\begin{deff}
The within-clusters sum of predictive influences is
\begin{equation} \label{objective}
C(\Theta,\mathcal{C})  = \sum_{k=1}^{K} \sum_{i \in \mathcal{C}_k} \lv \bm{\pi}_{\bm{x},\bm{y}}^{(k)}(\bm{x}_i,\bm{y}_i)\rv^2  ,
\end{equation}
where $\bm{\pi}_{\bm{x},\bm{y}}^{(k)}(\bm{x}_i,\bm{y}_i)$ is the predictive influence of a point $\{\bm{x}_i, \bm{y}_i \}$ under the $k^{th}$ TB-PLS model.
\end{deff} 

Clearly, when Eq. \eqref{objective} is minimised, each cluster consists of points that exert minimal predictive influence for that specific TB-PLS model, and therefore the overall prediction error is minimised. We refer to these optimal clusters as {\it predictive clusters}. If the true cluster assignments were known {\it a priori}, fitting these models and thus minimising the objective function would be trivial. However, since the true partitioning is unknown, there is no analytic solution to this problem, and we resort to an iterative algorithm that alternates between finding optimal cluster assignments and optimal model parameters. Specifically, the algorithm we suggest alternates between the following two steps:
\begin{enumerate}
\item
 Given $K$ TB-PLS models with parameters $\{\Theta_1,\ldots, \Theta_K\}$, and keeping these fixed, find the cluster assignments which solve
\begin{align}
\min_{\{ \mathcal{C}_1,...,\mathcal{C}_K \}} C(\Theta,\mathcal{C})  .
\label{obj_1}
\end{align}
\item
Given a set of cluster assignments, $\{ \mathcal{C}_1,...,\mathcal{C}_K \}$ and keeping these fixed, estimate the parameters of the $K$ predictive models which solve
\begin{align}
\min_{\{\Theta_1,\ldots, \Theta_K\}} C(\Theta,\mathcal{C}) .
\label{obj_2}
\end{align}
\end{enumerate}


We summarise the entire algorithm below.


\bigskip
{\bf Initialisation (I):} At iteration $\tau=0$, an initial, random partitioning of the data, $ \{ \mathcal{C}_1,...,\mathcal{C}_K \}$, is generated; both the TB-PLS models parameters, $\{\Theta_1,\ldots, \Theta_K\}$, and the predictive influences, $\bm{\pi}^{(k)}_{\bm{x},\bm{y}} (\bm{x}_i, \bm{y}_i)$, are computed for all $K$ clusters and $n$ observations. 

At each subsequent iteration $\tau=1, 2, ...$ the following two steps are performed in sequence until convergence:

{\bf Partitioning (P):} Keeping the model parameters fixed, the cluster assignments that minimise \eqref{obj_1} are obtained by assigning each point to the cluster for which its predictive influence is smallest, 
	\begin{equation}
	\mathcal{C}_k \leftarrow \left\{i: \min_k \lv \bm{\pi}^{(k)}_{\bm{x},\bm{y}} (\bm{x}_i, \bm{y}_i) \rv^2 \right\}.
	\label{eq_assignrule}
	\end{equation}

{\bf Estimation (E)}: Keeping the cluster allocations fixed, the parameters $\{\Theta_1,\ldots, \Theta_K\}$ that minimise \eqref{obj_2} are estimated using the data points $\{ \bm{x}_{i}, \bm{y}_{i} \}$ for all $i \in  \mathcal{C}_k$ and according to Eqs. \eqref{eq_pls_of}, \eqref{eq_q} and \eqref{eq_g}.

\bigskip


The computational cost of each iteration of the MVPP algorithm is of the order of computing an SVD in each cluster, $O(K(p^2q + q^2p))$. 

In order for the algorithm to converge to a local minimum we require that at each {\bf P} step and each {\bf E} step, the objective function must be decreasing. In Step {\bf  P} we assign observations to clusters based on the assignment rule \eqref{eq_assignrule} which minimises the predictive influence and so this decreases the objective function by definition. In Step {\bf E} we do not directly seek to minimise the predictive influence, instead we estimate parameters $\Theta^{new}_k$ in each cluster using TB-PLS. In order for these parameters to decrease the objective function it must be the case that they are closer to the optimal MVPP parameters, ${\Theta}^{*}_k$ than the parameters estimated using TB-PLS at the previous iteration ${\Theta}^{old}_k$. 

To see why this will be the case, we must consider what happens when points are reassigned to clusters. A large magnitude predictive influence is assigned to points which are influential under a given TB-PLS model. Therefore in Step {\bf E}, points which have been newly assigned to cluster $k$ will be influential under the TB-PLS model, $\Theta^{old}_k$ relative to other points in that cluster. If we estimate a new TB-PLS model, $\Theta^{new}_k$, these point will be assigned a smaller magnitude predictive influence and so the sum of square predictive influences within each cluster will be decreased. The algorithm converges to a local minimum of the objective function for any initial cluster configuration.

%

\subsection{Model selection} \label{sec_modsel}

Model selection in both clustering and TB-PLS are challenging problems which have previously only been considered separately. Within our framework, the PRESS statistic provides a robust method for efficiently evaluating the fit of the TB-PLS models to each cluster. A straightforward application of the PRESS allows us to identify the optimal number of clusters, $K$. We also apply a similar intuition to attempt to learn the number of latent factors of each TB-PLS model, $R_1,\ldots,R_K$.

Since our algorithm aims to recover predictive relationships on subsets of the data, the number of clusters is inherently linked to its predictive performance. If $K$ is estimated correctly, the resulting prediction error should be minimised since the correct model has been found. We therefore propose a method to select the number of clusters by minimising the out-of-sample prediction error which overcomes the issue of over-fitting as we increase $K$. The strategy consists in running the MVPP algorithm using values of $K$ between $1$ and some maximum value, $K_{max}$. We then select the value of $K$ for which  the mean PRESS value is minimised. This is possible due to our computationally efficient formulation of the PRESS for TB-PLS and the fact that we aim to recover clusters which are maximally predictive. The performance of this approach using simulated data is discussed in Section \ref{sec_no_clust}.

In the case where there is little noise in the data, the number of latent factors can be learned by simply evaluating the PRESS in each cluster at each iteration. Therefore, in the $k^{th}$ cluster, the value of $R_k$ is selected such that the PRESS is minimised. Since we select the value of $R_k$ which minimises the PRESS, this also guaranteed to decrease the objective function. However, as the amount of noise in the data increases, selecting each optimal $R_k$ value becomes a more difficult task due to the iterative nature of the algorithm. In this case, setting $R=1$ tends to capture the important predictive relationships which define the clusters whereas increasing each $R_k$ can actually be detrimental to clustering performance. This issue is discussed in Section \ref{sec_no_clust}. 

\section{Monte Carlo simulation procedures} \label{sec_simsettings}

\subsection{Overview}

In order to evaluate the performance of predictive partitioning and compare it to other multi-view clustering methods, we devise two different simulation settings which are designed to highlight situations both where current approaches to multi-view clustering are expected to succeed and fail.

 Commonly, clusters are considered to be formed by geometrically distinct groups of data points. This notion of geometric distance is also encountered implicitly in mixture models. Separation conditions have been developed for the exact recovery of mixtures of Gaussian distributions, for instance, for which the minimum required separation between means of the clusters is proportional to the cluster variances \cite{Kannan2008,Chaudhuri2009}.
	
In scenario A, we construct clusters according to the assumption that data points have a similar geometric structure under both views which should be recovered by existing multi-view clustering algorithms. We assess the performance as a function of the signal to noise ratio. As the level of noise is increased, the between-cluster separation diminishes to the point that all clusters are undetectable using a notion of geometric distance whereas a clustering approach based on predictive influence is expected to be more robust against noise. On the other hand, under scenario B the clustering of data points is not defined by geometric structures. We simulate data under cluster-wise regression models where the geometric structure is different in each view. In this situation, clustering based on geometric separation is expected to perform poorly regardless of the signal to noise ratio. In all of these settings we set the number of latent factors, $R=1$ and the number of clusters, $K=2$. A detailed description of these two settings is given below.

\subsection{Scenario A: geometric clusters}

The first simulation setting involves constructing $K$ geometric clusters (up until the addition of noise). We simulate each pairs of latent factors $\bm{t}^{(k)}$ and $\bm{s}^{(k)}$, with $k=1,\ldots,K$, from a bivariate normal distribution. Each $i=1,\ldots,n_k$ element, where $n_k=50$, is simulated as
 $( {t}_i^{(k)}, {s}_i^{(k)} )  \sim \mathcal{N}(\bm{\mu}_k , \bm{\Sigma}) ,$
where the means of the latent factors, $\bm{\mu}_{k}$ defines the separation between clusters. The covariance matrix is given by $\bm{\Sigma}$ having unit variances and off diagonal elements $\sigma_{ij}=0.9$. 
 
In order to induce a covariance structure in the $\bm{X}$ loadings, we first generate a vector $\bm{w}^{(k)}$ of length $p=200$ where each of the $m=1,\ldots,p$ elements is sampled from a uniform distribution
\begin{align*}
\bm{w}_m^{(k)} \sim \left\{ \begin{array}{ll} \text{Unif}[0,1], & \text{ if } m=1,\ldots,p/2 \\ \text{Unif}[1,2],  & \text{ if } m=p/2 + 1,\ldots,p \end{array} \right.
\end{align*}
The $p$ elements of the $\bm{X}$ loadings and the $q$ elements of the $\bm{Y}$ loadings are then simulated as
$\bm{u}^{(k)}  \sim \mathcal{N}\left(0,\bm{w}^{(k)}{\bm{w}^{(k)}}\tr\right)$, and  $\bm{v}^{(k)}  \sim \text{Unif}[0,1]$.
We then normalise the vectors so that $\lv\bm{u}^{(k)}\rv=\lv\bm{v}^{(k)}\rv=1$. Finally, for $K=2$, each pair of observations is generated from the TB-PLS model in the following way
\begin{align*}
\bm{x}_i = \left\{ \begin{array}{ll} {t}_i^{(1)} {\bm{u}^{(1)}}\tr+\bm{E}_{x,i}, & \text{ if } i\in \mathcal{C}_1 \\ {t}_i^{(2)} {\bm{u}^{(2)}}\tr+\bm{E}_{x,i},  & \text{ if } i\in \mathcal{C}_2 \end{array} \right.,
~~
\bm{y}_i = \left\{ \begin{array}{ll} {s}_i^{(1)} {\bm{v}^{(1)}}\tr+\bm{E}_{y,i}, & \text{ if } i\in \mathcal{C}_1 \\ {s}_i^{(2)} {\bm{v}^{(2)}}\tr+\bm{E}_{y,i},  & \text{ if } i\in \mathcal{C}_2 \end{array} \right.
\label{eq_gen_y}
\end{align*}
where each element of $\bm{E}_{x,i} \R ^{1\times p}$ and $\bm{E}_{y,i} \R ^{1\times q}$ are sampled i.i.d from a normal distribution, $\mathcal{N}(0,\sigma^2)$. The signal to noise ratio (SNR), and thus the geometric separation between clusters, is decreased by increasing $\sigma^2$.

\begin{figure}[htp] 
\centerline{\includegraphics[width=1.0\columnwidth]{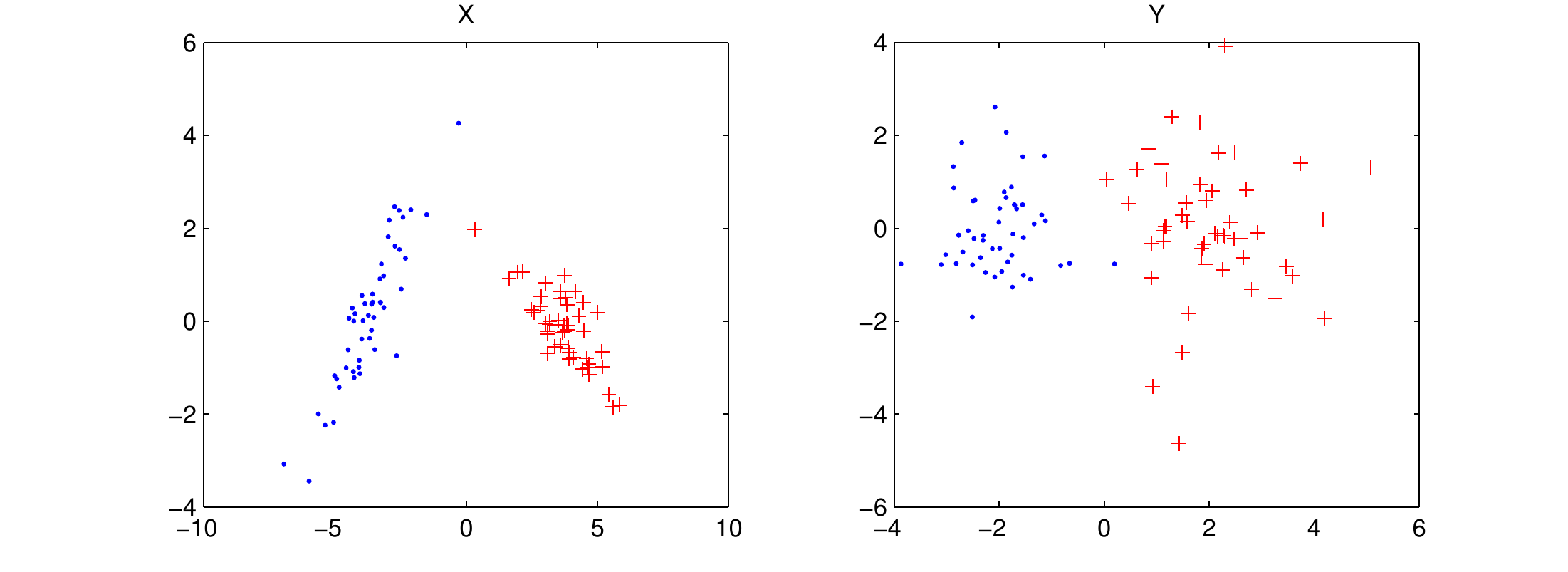}}
\caption{An example of data generated in scenario A where the  clusters are  ``geometric clusters'' i.e. the Euclidean distance between points within clusters is small compared to points between clusters. The predictors $\bm{X}\R^{100 \times 200}$ and response $\bm{Y}\R^{100 \times 200}$ have been plotted in the projected space.}
\label{fig_k2_lo}
\end{figure}

\marginpar{[{\bf FIGURE \ref{fig_k2_lo} AROUND HERE}]}

Figure \ref{fig_k2_lo} shows an example of data points generated under this simulation setting; the SNR is large and the geometric clusters are well separated. As the SNR decreases, the geometric clusters become less well separated and so this setting tests the suitability of the predictive influence for clustering when the data is noisy. 

\subsection{Scenario B: predictive clusters}

The second setting directly tests the predictive nature of the algorithm by breaking the link between geometric and predictive clusters. In this setting, the geometric position of the clusters in $\bm{X}$ and the predictive relationship between $\bm{X}$ and $\bm{Y}$ are no longer related. We start by constructing the data as in the previous section for $K=2$. However, we now split the first cluster in $\bm{X}$ space into three equal parts and translate each of the parts by a constant $c_1$. For all $i\in \mathcal{C}_1$
\[
	\bm{x}_{i} = \left\{ \begin{array}{lll} & \bm{x}_{i} + c_1 & \text{if } i=1,\ldots,n_k/3 \\
																&	 \bm{x}_{i}		 & \text{if } i=n_k/3+1,\ldots,2n_k/3 \\
																&	 \bm{x}_{i} - c_1 & \text{if } i=2n_k/3+1,\ldots,n_k  \end{array} \right.
\]  
We then split the second cluster in $\bm{X}$ space into two equal parts and perform a similar translation operation with a constant $c_2$. For all $i \in \mathcal{C}_2$
\[
	\bm{x}_{i} = \left\{ \begin{array}{lll} & \bm{x}_{i} + c_2 & \text{if } i=1,\ldots,n_k/2 \\
																 &	\bm{x}_{i}  & \text{if } i=n_k/2 + 1, \ldots, n_k  \end{array} \right.
\]  
The result is that there are now four distinct geometric clusters in $\bm{X}$ space but still only two clusters which are predictive of the points in $\bm{Y}$ space. Parametrising the data simulation procedure to depend on the constants $c_1$ and $c_2$ means that we can generate artificial datasets where one of the geometric clusters in $\mathcal{C}_1$ are geometrically much closer to $\mathcal{C}_2$ however the predictive relationship remains unchanged. We call these structures ``confounding clusters''.

\begin{figure} 
\centerline{\includegraphics[width=1.0\columnwidth]{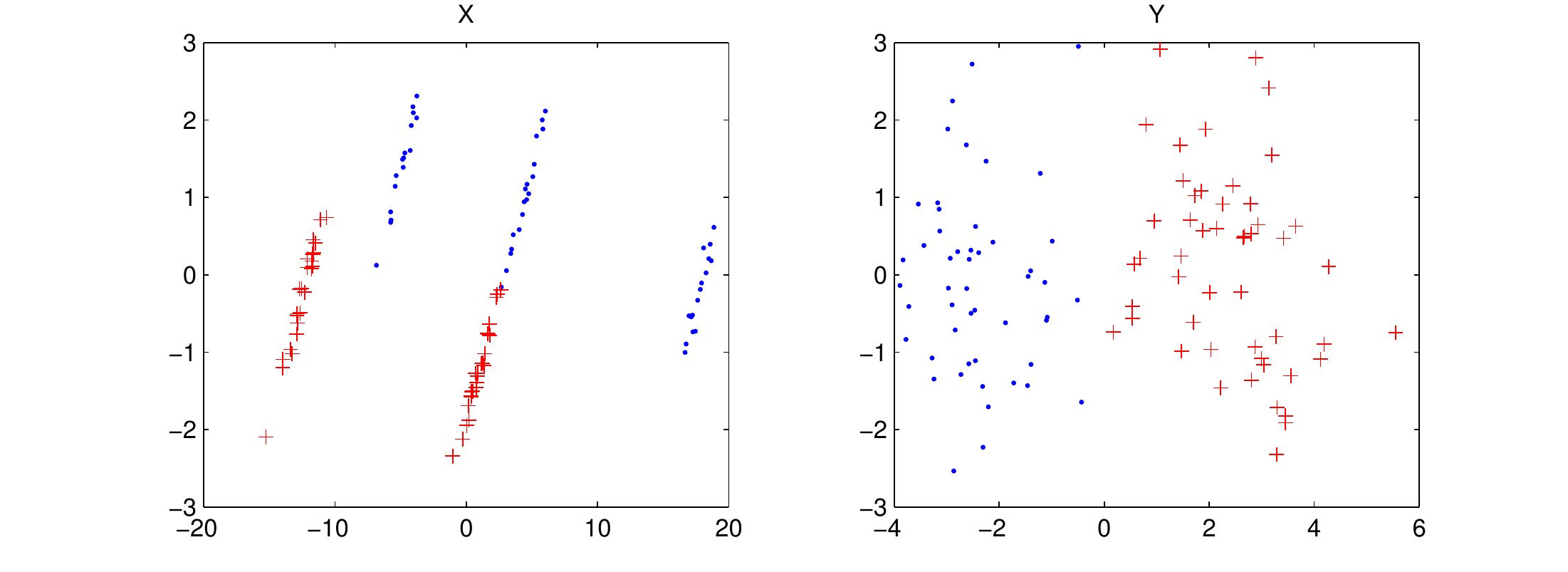}}
\caption{An example of data generated in scenario B. Points in the predictive clusters in $\bm{X}$ have been translated to create four clear geometric clusters. In this case, in the $\bm{X}$ view, the distance between cluster 2 (crosses) and two of the geometric clusters from cluster one (dots) is smaller than the distance between the points in cluster one. This implies that Euclidean distance based clustering will fail to recover the true clusters. The predictors $\bm{X}\R^{100 \times 200}$ and response $\bm{Y}\R^{100 \times 200}$ have been plotted in the projected space.}
\label{fig_split_lo}
\end{figure}

Figure \ref{fig_split_lo} shows an example of this simulation setting when the SNR is large. In this setting, noise is only added in the response which preserves the confounding geometric clusters in $\bm{X}$ but removes the separation between clusters in $\bm{Y}$. Therefore we expect methods which do not take into account predictive influence to fail to recover the true clusters and instead only recover the confounding geometric clusters.

\marginpar{[{\bf FIGURE \ref{fig_split_lo} AROUND HERE}]}

\section{Examples and simulation results} \label{sec_results}

\subsection{Influential observations} \label{sec_results_inf}

Initially we assess the ability of our criterion for detecting influential observations under a TB-PLS model, and demonstrate why using residuals only is unsuitable. For this assessment, we assume a homogeneous population consisting of bivariate points under each view, so $p=q=2$. We also assume that one latent factor only is needed to explain a large portion of covariance between the views.

\begin{figure}[htp]
\centering
\subfigure[Two-dimensional predictors and responses generated under the TB-PLS model. The influential observations are circled.]{
\includegraphics[width=1.0\columnwidth]{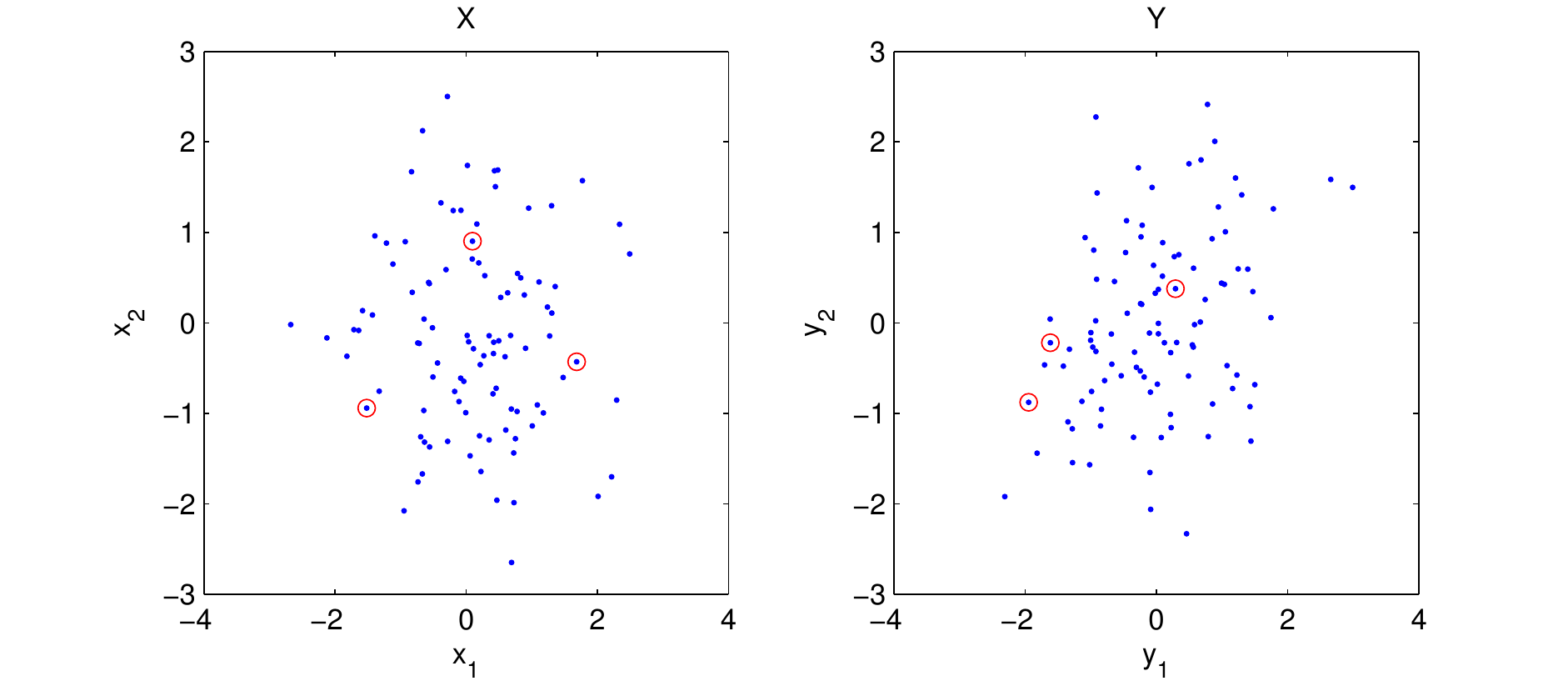}
\label{fig_outlier_data}
}
\subfigure[Two-dimensional predictors plotted against their corresponding magnitude residual error and predictive influence, respectively.]{
\includegraphics[width=1.0\columnwidth]{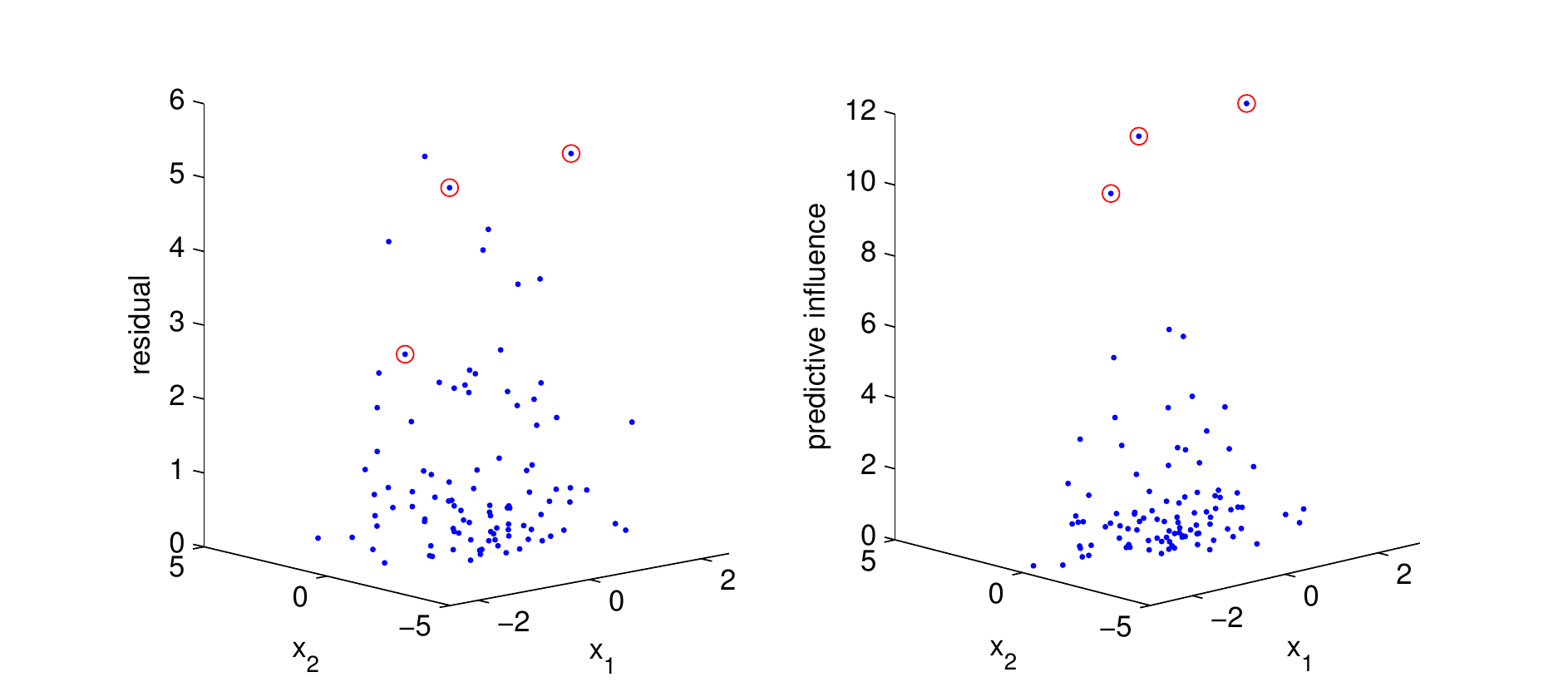}
\label{fig_outlier_pi}
}
\caption[]{\ref{fig_outlier_data} shows the two dimensional predictors, $\bm{X}$ and responses, $\bm{Y}$ with the influential observations circled. It is clear that the influential observations cannot be identified by simply examining these scatter plots. \ref{fig_outlier_pi} shows the magnitude residual (left-hand plot) and predictive influence (right-hand plot) for each observation in $\bm{X}$. The predictive influence of the influential observations is much larger than that of all other observations so that these points are clearly identified. The same degree of separation is not evident by examining the magnitude residual error.}
\label{fig_outlier}
\end{figure}

In order to generate data under the TB-PLS model, we first create the paired vectors $\{\bm{t},\bm{s}\}$ by simulating $n=100$ elements from a bivariate normal distribution with zero mean, unit variances and off diagonal elements $0.9$. The corresponding factor loadings $\bm{p}$ and $\bm{q}$ are simulated independently from a uniform distribution, Unif$(0,1)$. We then randomly select three observations in the $\bm{X}$ view and add standard Gaussian noise to each so that the between-view predictive relationship for those observations are perturbed. Figure \ref{fig_outlier_data} shows a plot of the predictors $\bm{X}$ and the responses $\bm{Y}$. The three influential observations are circled in each view. Since these observations are only different in terms of their predictive relationships, they are undetectable by visually exploring this scatter plot.

Using all $100$ points, we fit a TB-PLS model with $R=1$ and compute both the residual error and the predictive influence of each observation. In Figure \ref{fig_outlier_pi}, the observations in $\bm{X}$ are plotted against their corresponding residuals (shown in the left-hand plot) and predictive influences (shown in the right-hand plot). Since TB-PLS aims to minimise the residual error of all observations, including the influential observations results in a biased model fit; although the influential observations exhibit large residuals, this is not sufficient to distinguish them from non-influential observations. On the other hand, the predictive influence of each point is computed by implicitly performing leave-one-out cross validation and, as a consequence of this, the predictive influence of those points is larger than that of any of the other points. This simple example provides a clear indication that the influential observations can be identified by comparing the relative predictive influence between all points.


\marginpar{[{\bf FIGURE \ref{fig_outlier} AROUND HERE}]
}

We also perform a more systematic and realistic evaluation in higher dimensions. For this study, we simulate $300$ independent datasets, whereby each dataset has $p=q=200$, $n=100$ and three influential observations. We follow a similar simulation procedure as the one described before, and set $R=1$. Once the TB-PLS has been fit, all points are ranked in decreasing order, from those having the largest predictive influence and largest residual. We then select the first top $m$ ranked observations (with $m=1, \ldots, n$) and define a true positive as any truly influential observation that is among the selected ones; all other observations among those $m$ are considered false positives. 

\begin{figure}[htp] 
\centerline{\includegraphics[width=0.55\columnwidth]{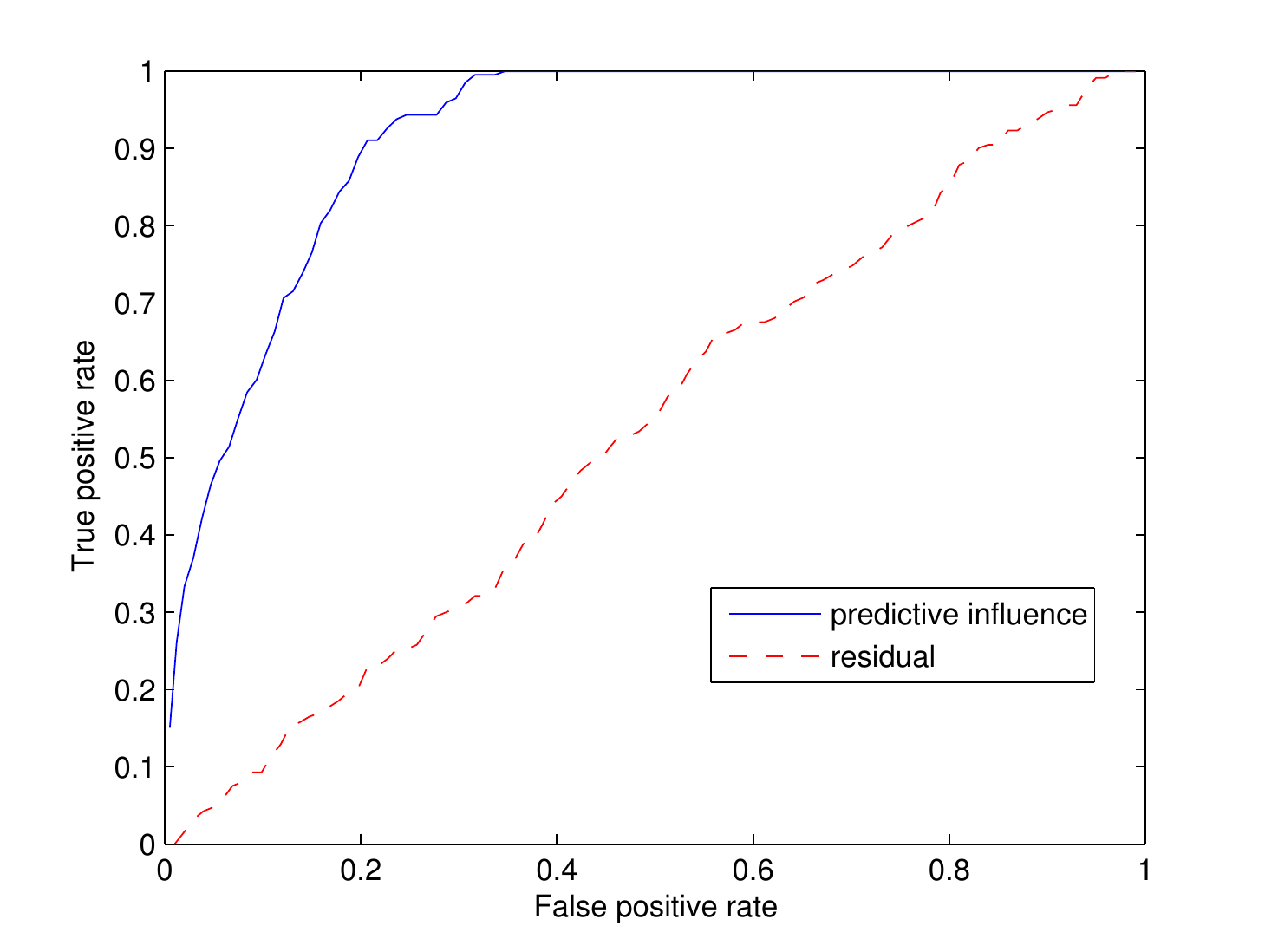}}
\caption{Receiver operating characteristic (ROC) curve which compares the ability to detect outliers of the predictive influence and the residual in high dimensions $(p=q=200)$. The results are averaged over 300 Monte Carlo simulations. Using the predictive influence to detect influential observations consistently identifies more true positives for a given false positive rate than using the residual. The predictive influence detects all influential observations with a false positive rate of $0.34$ whereas the residual consistently identifies almost as many false positives as true positives.}
\label{fig_roc}
\end{figure}

Figure \ref{fig_roc} compares the receiver operating characteristic (ROC) curve obtained using the predictive influence and the residual error for this task. This figure shows that the predictive influence consistently identifies the true influential observations with fewer false positives than when the residual is used.
 This suggests that using the residuals to detect influential observations in high dimensions is approximately equivalent to a random guess, and clearly demonstrates the superiority of the proposed predictive influence measure for this task.

\marginpar{[{\bf FIGURE \ref{fig_roc} AROUND HERE}]}

\subsection{Confounding geometric structures}

Here we revisit the illustrative example described in Section \ref{sec_heterogeneous}. The predictors shown in Figure \ref{fig_subspace} consist of three-dimensional points sampled uniformly along a line and a plane, and these two subspaces intersect. The response consists of a noisy linear combination of the points in each cluster. Using the same simulated data, we can explore the performance of both our MVPP algorithm and a different multi-view clustering algorithm, MV-CCA \cite{Chaudhuri2009}. MV-CCA fits a single, global CCA model which assumes all points belong to the same low dimensional subspace and that clusters are well separated geometrically in this subspace.

\begin{figure}[htp] 
\subfigure[Clustering using MV-CCA]{
\centerline{\includegraphics[width=1\columnwidth]{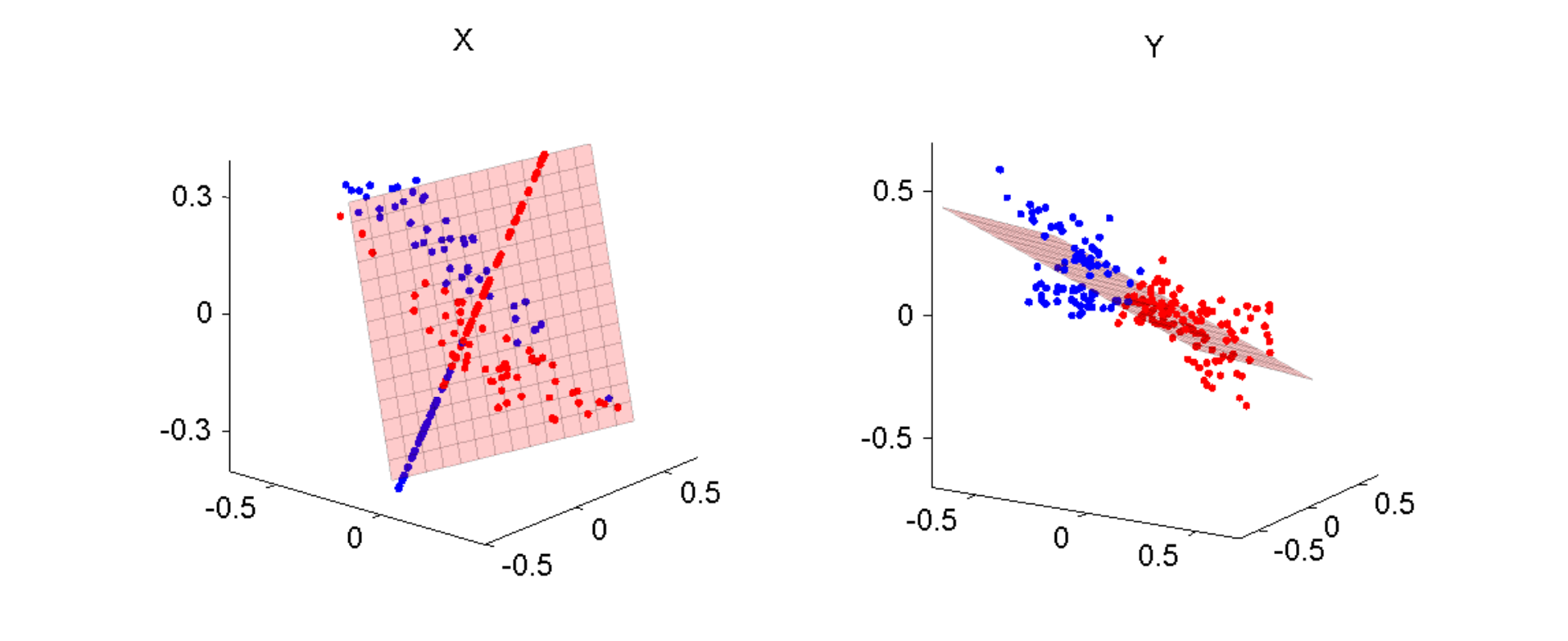}}
\label{fig:subfig3}}
\subfigure[Clustering using MVPP]{
\centerline{\includegraphics[width=1\columnwidth]{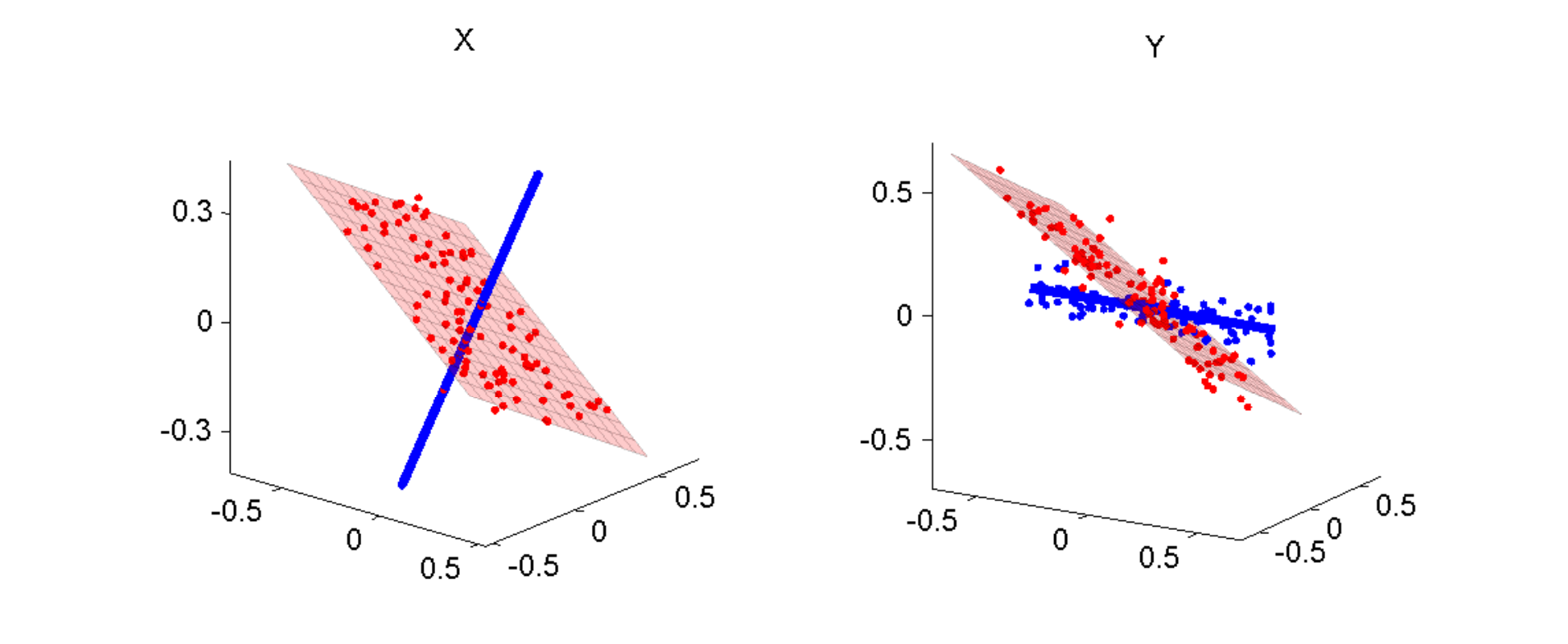}}
\label{fig:subfig4}}
\caption{Plot \subref{fig:subfig3} shows the result of clustering the example dataset introduced in Figure \ref{fig_subspace} using the MV-CCA method. It can be seen that MV-CCA fits a single plane to the data and assigns points to clusters based on geometric distances between points on that plane so the resulting clustering is incorrect.  Plot \subref{fig:subfig4} shows the result of clustering using the MVPP  algorithm which models the predictive relationship within each cluster. As a result, the true subspaces and cluster assignments are recovered.}
\label{fig_subspace2}
\end{figure}

Figure \ref{fig_subspace2} shows the clustering results on this example dataset using both MV-CCA and MVPP. The result of clustering using MV-CCA shown in Figure \ref{fig:subfig3} highlights the weaknesses of using a global, geometric distance-based method since the existence of clusters is only apparent if local latent factors are estimated using the correct subset of data. MV-CCA fits a single plane to the data which is similar to the one estimated by a global TB-PLS model, as in  Figure \ref{fig:subfig2}. The points are then clustered based on their geometric separation on that plane which results in an incorrect cluster allocation. 

In comparison, Figure \ref{fig:subfig4} shows the result of clustering with MVPP showing how the ability to recover the true clusters, and therefore deal with the confounding geometric structures, by inferring the true underlying predictive models. Moreover, since the noise in the data low, in this example we are able to let MVPP learn the true number of latent factors in each cluster using the procedure described in Section \ref{sec_modsel}.

\marginpar{[{\bf FIGURE \ref{fig_subspace2} AROUND HERE}]}

\subsection{Clustering performance}

Using data simulated under scenarios A and B, we assess the mean clustering and predictive performance of the MVPP algorithm in comparison to some multi-view clustering algorithms over 200 Monte Carlo simulations. In each simulation, the latent factors, loadings and noise are randomly generated as described in section \ref{sec_simsettings}. We also examine issues relating to model selection in the MVPP algorithm.

\begin{figure}[htp] 
\centerline{\includegraphics[width=0.6\columnwidth]{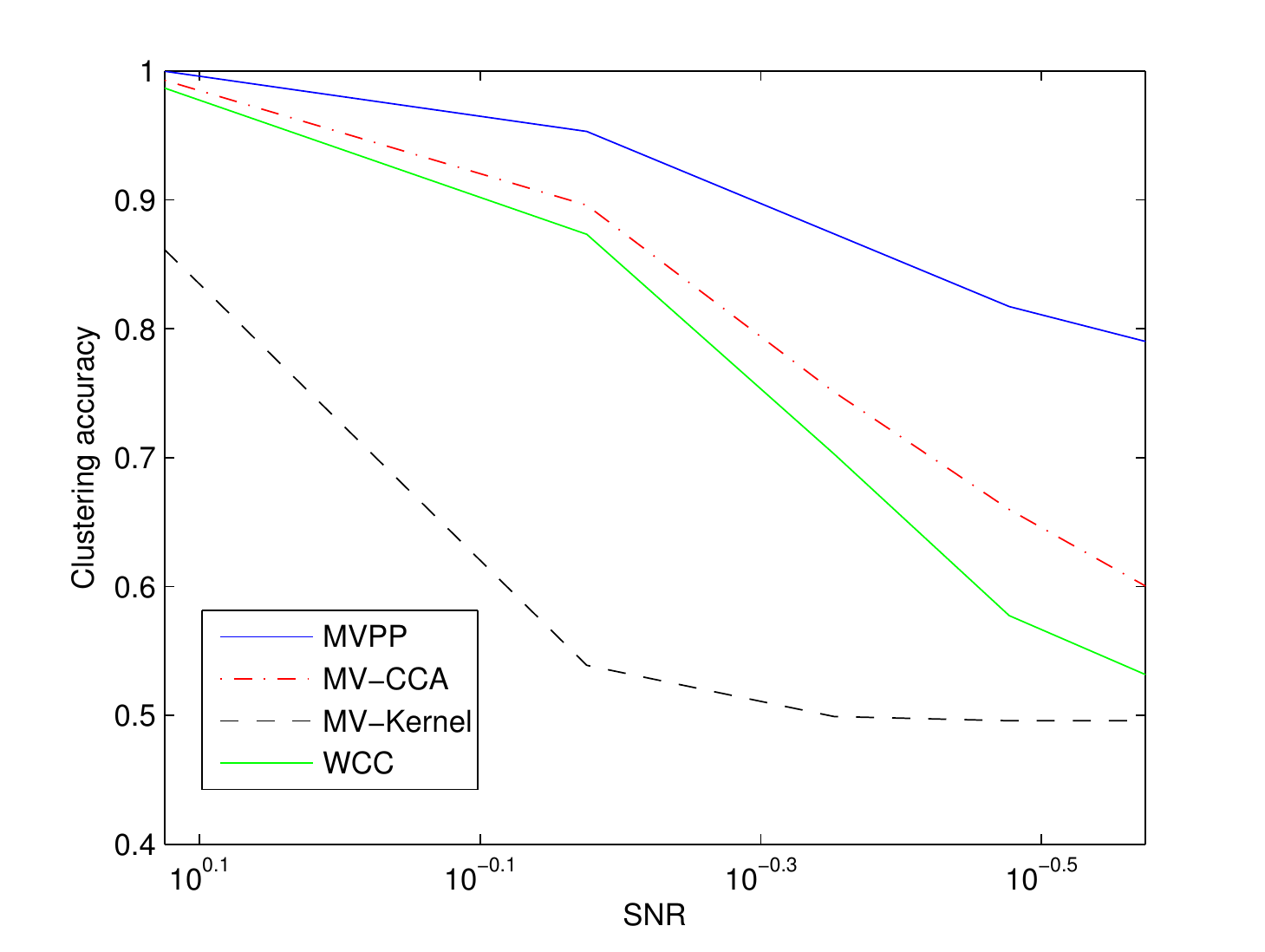}}
\caption{Comparing the mean clustering accuracy of different methods for $K=2$ in simulation setting A over 200 Monte Carlo simulations. When the SNR is high, MVPP achieves maximum accuracy and as the noise increases, the decrease in performance is small relative to the other methods. }
\label{fig_clusterK2}
\end{figure}

\marginpar{[{\bf FIGURE \ref{fig_clusterK2} AROUND HERE}]}

Figure \ref{fig_clusterK2} shows the result of the comparison of clustering accuracy between methods when $K=2$ in scenario A. A SNR of $10^{0.1}$ indicates that signal variance is approximately $1.3$ times that of the noise variance and so the clusters in both views are well separated whereas a SNR of $10^{-0.5}$ indicates that the clusters overlap almost completely. It can be seen that when the noise level is low, MVPP is able to correctly recover the true clusters. As the noise increases, and the geometric separation between clusters is removed, the clustering accuracy of the competing methods decreases at a faster rate than MVPP. 

Since MV-CCA assumes that the clusters are well separated geometrically, as the noise increases the estimated latent factor is biased which decreases the separation between the clusters. Another reason for the difference in performance between MV-CCA and MVPP lies with how the multiple views are used for clustering. Although MV-CCA clustering derives a low dimensional representation of the data using both views, the actual clustering is performed using the latent factors of only one view. MVPP considers the important predictive contribution from both views in constructing the predictive influences and so clustering occurs jointly between the views.

The MV-kernel method \cite{Sa2010} relies on the Euclidean distance between points in constructing the similarity matrix. This method works well only when the clusters are well separated in each view.  Computing the Euclidean distance between points in high dimensions before performing dimensionality reduction means that the MV-kernel method is affected by the curse of dimensionality. As such, its performance degrades rapidly as the SNR decreases.

WCC \cite{Li} clusters each view separately using $K$-means and combines the partitions to obtain a consensus. Since it does not take into account the relationship between the two views, when the data is noisy this can result in two extremely different partitions being recovered in each view and therefore a poor consensus clustering.

\begin{figure}[htp] 
\centerline{\includegraphics[width=0.6\columnwidth]{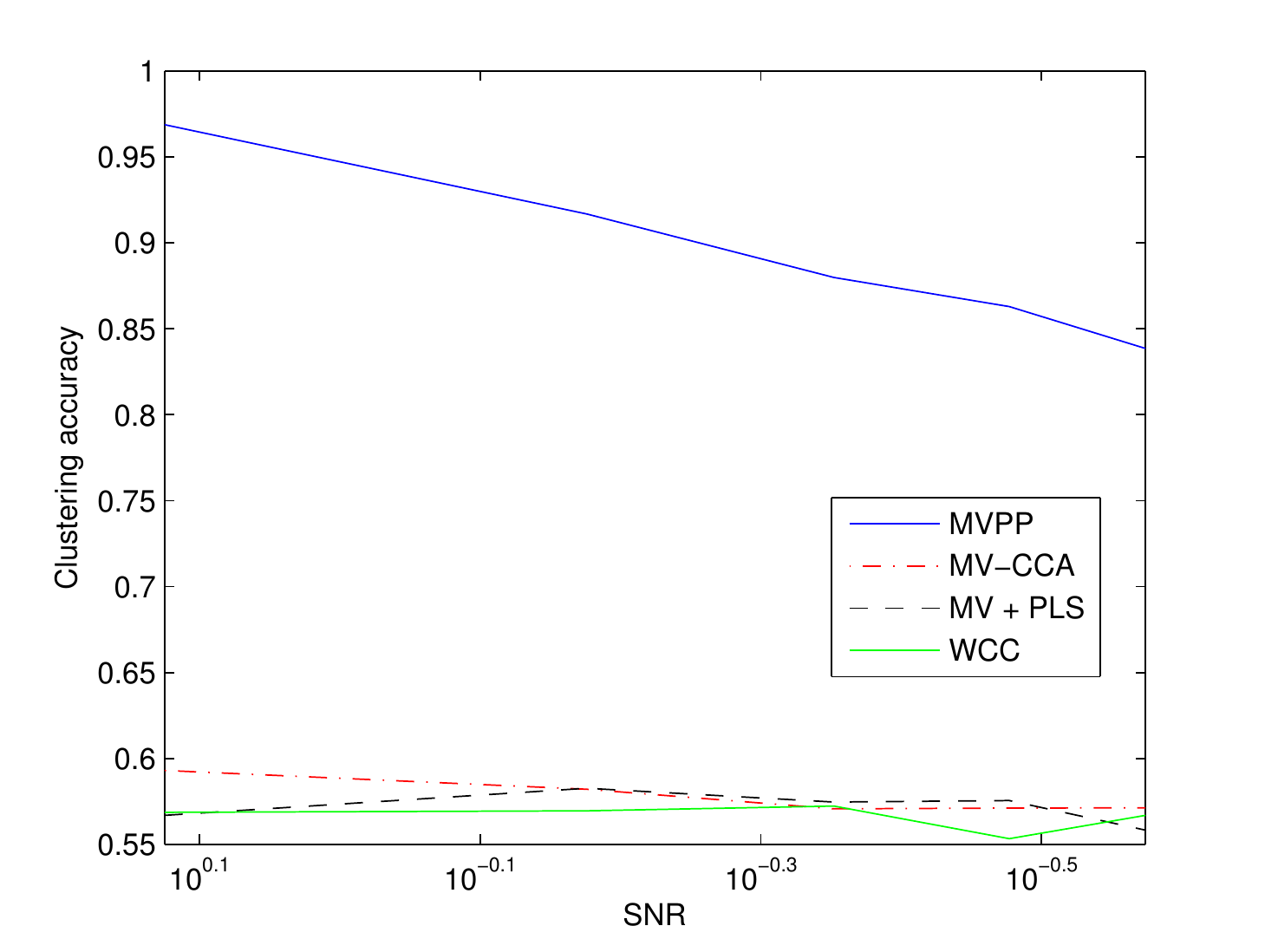}}
\caption{Comparing the mean clustering accuracy in simulation setting B over 200 Monte Carlo simulations. MVPP achieves a high clustering accuracy for all levels of noise whereas the competing methods perform poorly even when the SNR is large, since they recover clusters based on the confounding geometric structure in the $X$ view. }
\label{fig_cluster_hard}
\end{figure}

\marginpar{[{\bf FIGURE \ref{fig_cluster_hard} AROUND HERE}]}

Figure \ref{fig_cluster_hard} shows the result of the comparison between methods in scenario B. It can be seen that MVPP consistently clusters the observations correctly in this challenging setting and is extremely robust to noise due to the implicit use of cross-validation. Since none of the other methods takes into account the predictive relationship between the clusters and instead only find geometric clusters, they all consistently fail to identify the true clusters. The similar performance for low levels of noise corresponds to these methods consistently misclustering the points based on their geometric position. As the noise increases, the performance of  WCC, MV-CCA and MV-kernel remains fairly constant. This confirms that these methods are not correctly utilising the important information in the second view of data even when the predictive clusters in the response are well separated.

\subsection{Predictive performance}

Since only MVPP considers the predictive performance of the clustering by evaluating the PRESS error in each cluster, in order to test the predictive performance of the competing multi-view clustering algorithms we must perform clustering and prediction in two steps.
Therefore we first perform clustering with each of the methods on the full dataset and then train a TB-PLS model in each of the obtained clusters. We then test the predictive ability by evaluating the leave-one-out cross validation error within each cluster. For comparison, we also evaluate the LOOCV error of a global TB-PLS model which we fit using all of the data.  

\begin{figure}[htp] 
\centerline{\includegraphics[width=0.6\columnwidth]{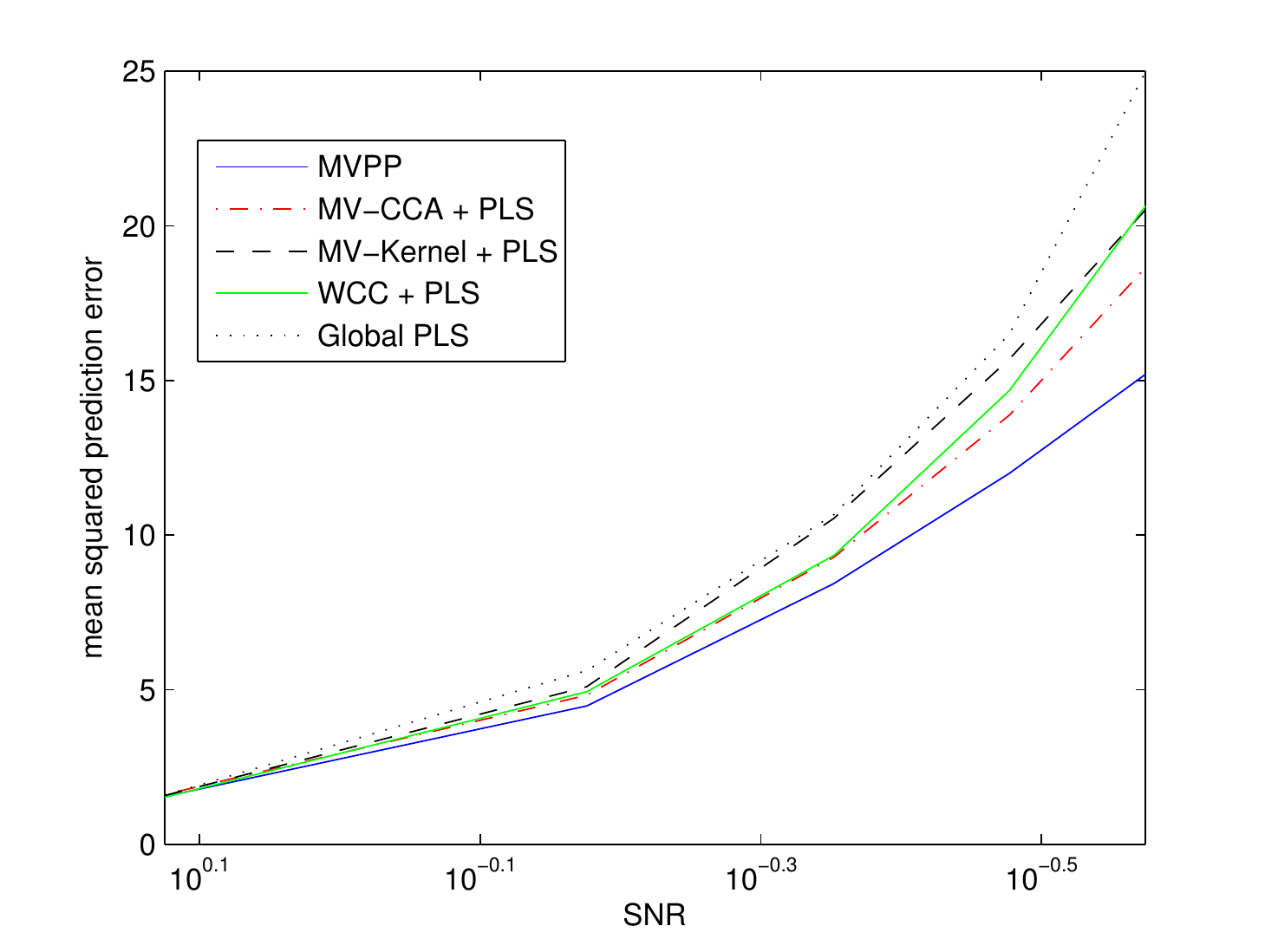}}
\caption{Comparing the mean leave-one-out prediction error over 200 Monte Carlo simulations of the clusters obtained by different methods for $K=2$ in simulation setting A. MVPP consistently achieves the lowest prediction error of the multi-view clustering methods due to the clusters being selected based on their predictive ability. Similarly to the clustering performance, as the noise increases the relative difference between MVPP and the other methods also increases. It can be seen that all clustering methods achieve better prediction than a global PLS model.}
\label{fig_predictK2}
\end{figure}

\marginpar{[{\bf FIGURE \ref{fig_predictK2} AROUND HERE}]}

Figure \ref{fig_predictK2} shows the result of predictive performances under scenario A. This figure shows that MVPP achieves the lowest prediction error amongst the multi-view clustering methods. This is to be expected since the clusters are specifically obtained such that they are maximally predictive through implicit cross validation.  The prediction error of the competing multi-view methods is larger than MVPP which indicates that these methods are really not selecting the truly predictive clusters. As the noise increases, the prediction performance of all methods decreases however as MVPP is more robust to noise than the competing methods, its relative decrease in performance is smaller. 
It can be noted that for low levels of noise the global predictive model performs worst of all. This further supports the notion of attempting to uncover locally predictive models within the data.

\begin{figure}[htp] 
\centerline{\includegraphics[width=0.6\columnwidth]{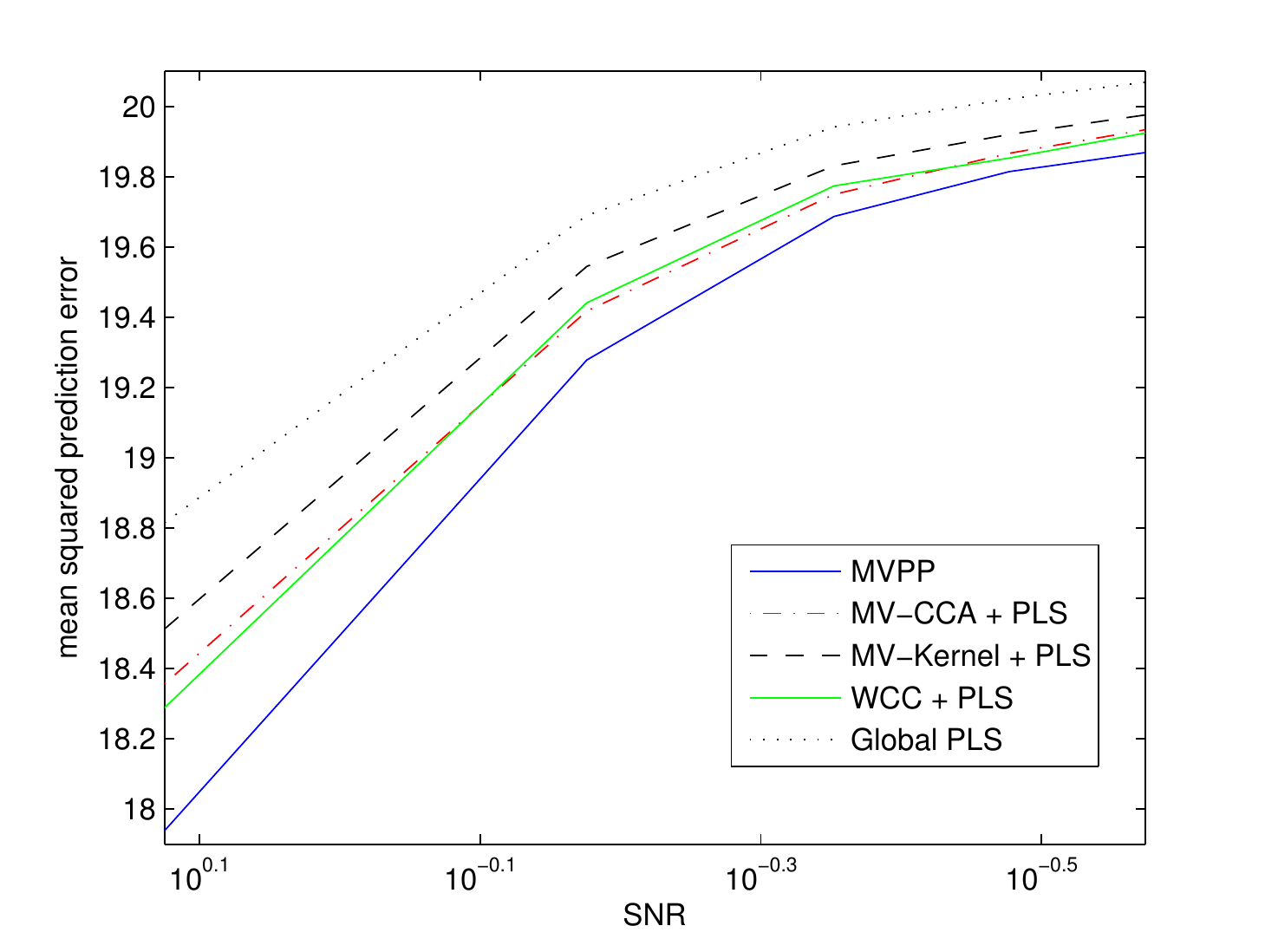}}
\caption{Comparing the mean leave-one-out prediction error of the clusters obtained in simulation setting B over 200 Monte Carlo simulations. MVPP achieves the best prediction performance of the multi-view clustering methods. Since as noise increases, the relative clustering clustering performance between MVPP and the competing methods decreases, this relative predictive performance of MVPP also decreases. Again, global PLS achieves the worst prediction accuracy of all methods.}
\label{fig_predict_hard}
\end{figure}

\marginpar{[{\bf FIGURE \ref{fig_predict_hard} AROUND HERE}]}

Figure \ref{fig_predict_hard} shows the prediction performance in scenario B. Since MVPP is able to accurately recover the predictive clusters, it displays the lowest prediction error amongst the multi-view clustering methods. As noted above, the other multi-view clustering methods only recover the geometric clusters and so their prediction performance is worse. The relative performance difference between competing methods stays similar as noise increases however, since MVPP is affected by noise in $\bm{Y}$, its predictive performance decreases relative to the other methods. 


\subsection{Model selection}

\label{sec_no_clust}
\begin{figure}[htp] 
\centerline{\includegraphics[width=1\columnwidth]{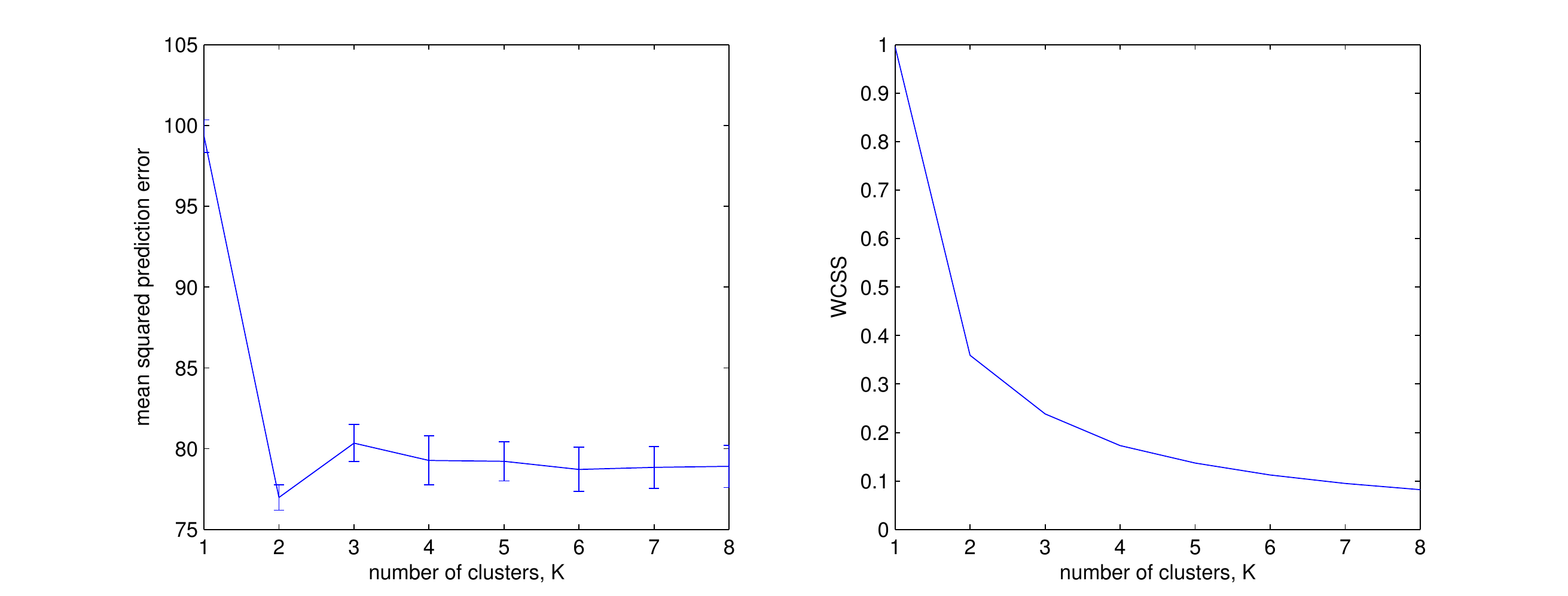}}
\caption{Comparing the prediction error with the objective function for different values of $K$ in the first simulation setting where the true value of $K=2$. It can be seen that as $K$ increases the global minimum of the PRESS occurs when $K=2$, whereas the objective function decreases monotonically as it begins to overfit the data. The error bars also show that the standard deviation of the PRESS is smallest when $K=2$. This allows us to use the prediction error to select the true number of clusters.}
\label{fig_choosek}
\end{figure}

%
The ability of MVPP to learn the true number of clusters in the data is assessed using the procedure in Section \ref{sec_modsel}. In this experiment, the data was simulated under setting A and the true number of clusters was set as $K=2$. Figure \ref{fig_choosek} shows a comparison between the PRESS prediction error and the objective function for different values of $K$ over 200 Monte Carlo simulations.  As expected, the objective function decreases monotonically as $K$ is increased whereas the PRESS exhibits a global minimum at $K=2$. 

\marginpar{[{\bf FIGURE \ref{fig_choosek} AROUND HERE}]}

In the above simulation settings, the number of latent factors was fixed to be $R=1$.
According to the TB-PLS model in Section \ref{sec_pls} the first latent factor is the linear combination of each of the views which explains maximal covariance between $\bm{X}$ and $\bm{Y}$. Therefore, the first latent factor is the most important for prediction. Each successive latent factor explains a decreasing amount of the covariance between the views and so contributes less to the predictive relationship.

\begin{figure}[htp] 
\centerline{\includegraphics[width=0.6\columnwidth]{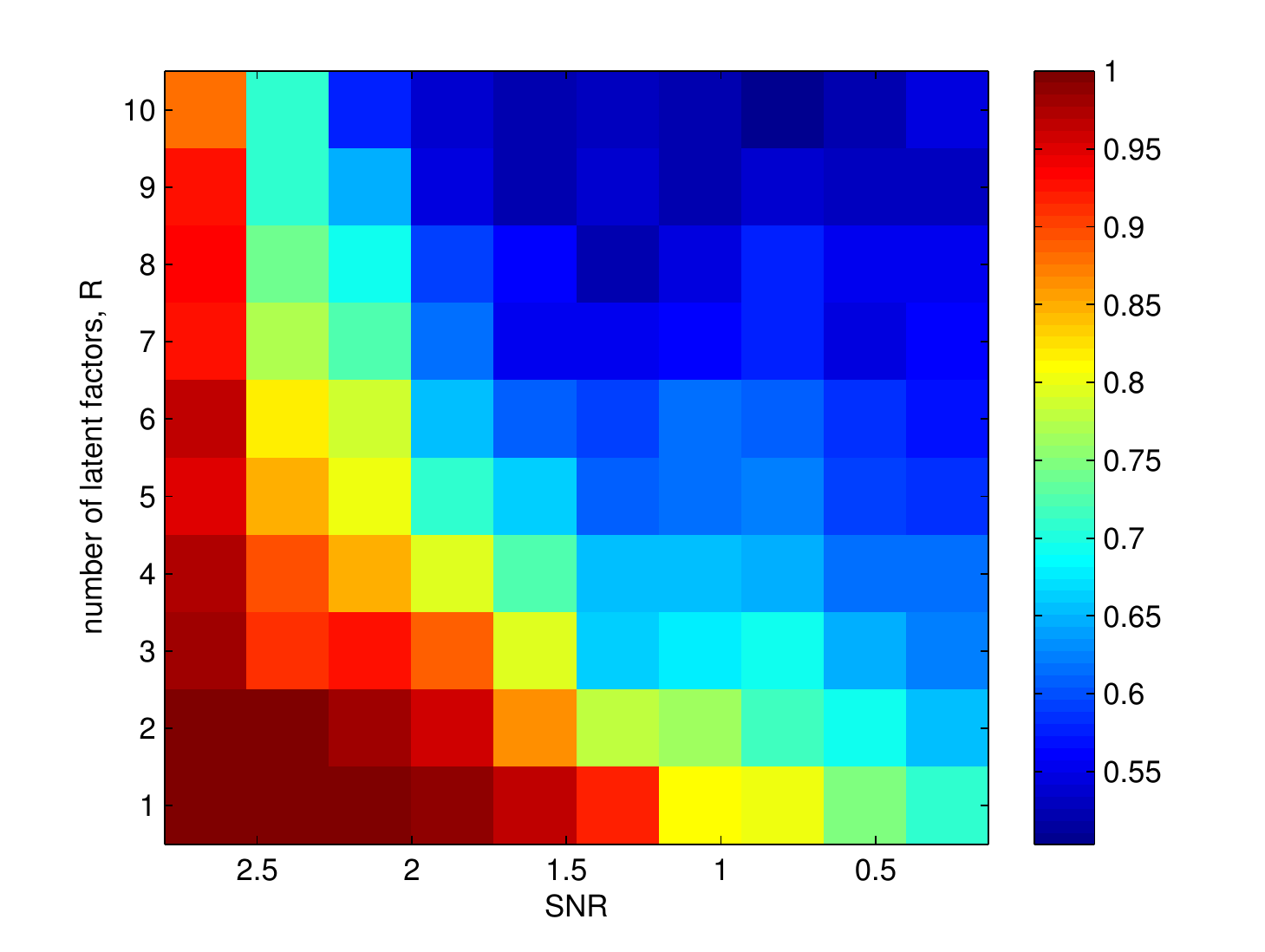}}
\caption{The effect of the number of latent factors, $R$ on the clustering accuracy. For low levels of noise, increasing $R$ has little effect on the clustering accuracy. However, as the noise increases, it can be seen that the first latent factor explains all of the signal in the data and increasing $R$ has a detrimental effect on the clustering accuracy. }
\label{fig_chooser}
\end{figure}

Figure \ref{fig_chooser} shows the effect of the number of latent factors, $R$ on the clustering accuracy of MVPP in scenario A. It can be seen that for low levels of noise, when the clusters are well separated, increasing $R$ has little effect on the clustering accuracy. As the noise increases, the first latent factor appears to capture all of the important predictive relationships in the data whereas subsequent latent factors only fit the noise which causes a detrimental effect on the clustering accuracy as more latent factors are added.


\marginpar{[{\bf FIGURE \ref{fig_chooser} AROUND HERE}]}

\section{An applications to web data} \label{sec_results2}

\subsection{Data description}

The proposed MVPP method, as well as alternative multi-view clustering algorithms, have been tested on two real world datasets. The first is the WebKB\footnote{http://www.cs.cmu.edu/~webkb} dataset which consists of a collection of interconnected web pages taken from the computer science departments of four universities: Cornell, Texas, Washington and Wisconsin. This dataset is commonly used to test multi-view clustering algorithms, and therefore provides an ideal benchmark \cite{Bickel2004,Hou2010,Sa2010}. We treat each web page as an observation, where the predictor vector, $\bm{x}_i$ is the page text and the response vector, $\bm{y}_i$ is the text of the hyperlinks pointing to that page. The dimensions of these vectors for each university is given in Table \ref{tab_webkb}. Both views of the pages consist of a bag of words representation of the important terms where the stop words have been removed. Each word has been normalised according to its frequency in each page and the inverse of the frequency at which pages containing that word occur (term frequency-inverse document frequency normalisation). 

There are two separate problems associated with the WebKB dataset. The first problem, which we denote WebKB-2, involves clustering the pages into two groups consisting of ``course'' and ``non-course'' related pages respectively. The second problem, WebKB-4, involves clustering the pages into four groups consisting of ``course'', ``student'', ``staff'' and ``faculty'' related pages. It is known that a predictive relationship exists between views \cite{Tian2006} and so we expect the results obtained by MVPP to reflect the ability to exploit that relationship in order to correctly identify clusters. 

\marginpar{[{\bf TABLE \ref{tab_webkb} AROUND HERE}]}

We also evaluate the clustering and prediction performance of MVPP and competing methods on a second benchmark dataset, the Citeseer dataset \cite{Bickel2005}, which consists of scientific publications $(n=3312)$ belonging to one of six classes of approximately equal sizes. The predictor view, $\bm{x}_i$, consists of a bag of words representation of the text of each publication in the same form as the WebKB dataset $(p=3703)$. We perform two different analyses: in the first one, the response view $\bm{y}_i$ comprises of a  binary vector of the incoming references between a paper and the other publications in the dataset ($q=2316$); in the second, the response view comprises of a  binary vector of the outgoing references from each paper ($q=1960$).

\subsection{Experimental results}

For the WebKB-2 clustering problem there are two true clusters of approximately equal size. We again compare MVPP with the WCC, MV-CCA and MV-Kernel clustering methods. For each method, we then evaluate the leave-one-out prediction error for the previously recovered clusterings. We also evaluate the leave-one-out prediction error for global PLS which has been estimated using all the data.

\marginpar{[{\bf TABLE \ref{tab_webkb_2} AROUND HERE}]}

Table \ref{tab_webkb_2} shows the results of clustering and prediction on the WebKB-2 dataset. In all cases, MVPP achieves almost 100\% clustering accuracy whereas the other methods achieve between $50-87\%$ accuracy which suggests that there is a predictive relationship between the text view of the webpage and the incoming links which MVPP is able to exploit to recover the true clusterings. MVPP also achieves a much lower prediction error than the other clustering methods which vary widely. This suggests that since the dimensionality of the problem is large, a small error in cluster assignment can lead to fitting a poor predictive model. 
 
\marginpar{[{\bf TABLE \ref{tab_webkb_4} AROUND HERE}]}

For the WebKB-4 clustering problem there are four true clusters where one of the clusters is much larger than the others. This poses a particularly challenging scenario since K-means based techniques favour clusters which are of a similar size. Table \ref{tab_webkb_4} details the results on this dataset. Again, in all cases, MVPP achieves the highest clustering accuracy. In this dataset, the clustering accuracy for MVPP is approximately 15\% lower than for $K=2$ due to the irregular cluster sizes and the poorer separation between clusters. The other methods also generally achieve poorer clustering accuracy however the relative decrease is not as large. Similarly for the previous dataset, the better clustering performance of the multi-view methods does not necessarily achieve better prediction performance. Despite achieving a relatively poorer clustering accuracy, fitting four clusters instead of two greatly improves the prediction performance of all clustering methods. 

\marginpar{[{\bf TABLE \ref{tab_citeseer_6} AROUND HERE}]}

Table \ref{tab_citeseer_6} shows the results for clustering and prediction using the Citeseer dataset. It can be seen that in both configurations, MVPP achieves the highest clustering accuracy although the relative difference is not as large as for the WebKB dataset. In this case, MVPP achieves the lowest prediction error of all methods. The large variance in prediction error between the multi-view clustering methods despite their similar clustering accuracy again suggests that incorrectly clustering observations can severely affect the prediction performance due to the high dimensionality of the data. 

\section{Conclusions} \label{sec_conc}

In this work, we have considered the increasingly popular situation in machine learning 
of identifying clusters in data by combining information from multiple views. We have highlighted some cases where the notion of a predictive cluster can better uncover the true partitioning in the data. In order to exploit this, our work consolidates the notion of predictive and cluster analysis which were previously mostly considered separately in the multi-view learning literature. 

In order to identify the true predictive models in the data, we have developed a novel method for assessing the predictive influence of observations under a TB-PLS model. We then perform multi-view clustering based on grouping together observations which are similarly important for prediction. The resulting algorithm, MVPP, is evaluated on data simulated under the TB-PLS model such that the true clusters are predictive rather than geometric. The results demonstrate how geometric distance based multi-view clustering methods are unable to uncover the true partitions even if those methods explicitly assume the data is constructed using latent factors. On the other hand, MVPP is able to uncover the true clusters in the data to a great degree of accuracy even in the presence of noise and confounding geometric structure. Furthermore, the clusters obtained by MVPP provide the basis of a better predictive model than the clusters obtained by the competing methods. An application to real web page and academic paper data show similar results. 

The computational complexity of MVPP is at least $K$-times more expensive compared with the other CCA-based multi-view clustering algorithms. This computational cost in incurred due to the need to iteratively fit a predictive model in each cluster which can be expensive when the dimensionality of the data is high. However, as shown by our results on simulated and real datasets, it appears that such a strategy is necessary in order to recover an accurate partitioning of the data.

Determining an initial partitioning which performs better than a random initialisation is a difficult problem since identifying the local predictive relationships \emph{a priori} is not always possible using global methods. An obvious choice would be to initialise the algorithm using the results from the MV-CCA method. However, it can be seen that in certain situations (such as scenario B), MV-CCA results in a poor cluster assignment which may result in the MVPP algorithm getting stuck in a poor local minimum.

We have also attempted to unify the difficult issues of model selection in clustering and TB-PLS which have previously only been considered separately. We have shown that our prediction based clustering criterion can be used to learn the true number of clusters. However, we have also seen that learning the number of latent factors in each of the TB-PLS models remains a difficult problem due to the effects of noise and the iterative nature of the algorithm.

The idea of multi-view clustering based on prediction has not been explored before in the literature, but there are examples of clustering using mixtures of linear regression models in which the response is univariate  \cite{Bishop2006}. 
However, it is well known that the least squares solution is prone to over-fitting and does not represent the true predictive relationship inherent between the views. Furthermore, the least squares regression applies only to a univariate response variable, and is not suitable for situations where the response is high dimensional. 



A possible extension to the MVPP method for high dimensional and noisy data is to apply an additional constraint on the $\ell_1$ norm of the TB-PLS weights estimated in Eq. \eqref{eq_pls_of}. Such a constraint induces sparsity in the TB-PLS solution such that only a small number of variables contribute to the predictive relationship between the views. This can be achieved, for example, using the Sparse PLS method of \cite{McWilliams2010}. However, this requires the specification of additional tuning parameters which cannot be easily learned from the data.

\appendix

\section{PLS PRESS}
For one latent factor we can write the $i^{th}$ leave one out error as
$\bm{e}_{-i} =  \bm{y}_i - \bm{x}_i \bm{\beta}_{-i}$ ,
where $\bm{\beta}_{-i} $ is estimated using all but the $i^{th}$ observation. Since $\bm{\beta}=\bm{u} g \bm{q}\tr$, we can write
$\bm{e}_{-i} =  \bm{y}_i - \bm{x}_{i} \bm{u}_{-i} g_{-i} \bm{q}_{-i}\tr  $.
The difference between the singular vectors estimated using all the data and the leave-one-out estimate, $\lv \bm{u} - \bm{u}_{-i}\rv$ is of order $O\left(\sqrt{\frac{\log (n-1)}{n-1}}\right)$ \cite{McWilliams2010a} so that if $n$ is large, we can write
$\bm{e}_{-i} =  \bm{y}_i - \bm{x}_i \bm{u} g_{-i} \bm{q}_{-i}\tr $.

Using the matrix inversion lemma, we can obtain recursive update equations for $g_{-i}$ which only depends on $g$ and does not require an explicit leave-one-out step in the following way 
\begin{align}{g}_{-i}  
						=  { g} - \frac{(s_i - t_i {g}) (\bm{t}\tr \bm{t})^{-1} t_i}{1 - t_i^2} ,\end{align}
 where the expression for $g$ is given by Equation \eqref{eq_g}. In the same way we derive the following expression
$\bm{q}_{-i}  =  \bm{q} - \frac{( \bm{y_i} - s_i \bm{q}\tr ) (\bm{s}\tr \bm{s})^{-1} s_i}{1 - s_i^2}$ ,
where the expression for $\bm{q}$ is given by Equation \eqref{eq_q}. Equation \eqref{eq_press_refac} is then obtained by using these values for $g_{-i}$ and $\bm{q}_{-i}$ in $\bm{\beta}_{-i}=\bm{u}g_{-i}\bm{q}_{-i}\tr$ and simplifying.

\section{Predictive influence}
\label{sec_appendixB}
Taking the partial derivative of the PRESS function, $J$ with respect to $\bm{x}_i$
\begin{align*}
\frac{\partial J}{\partial \bm{x}_i}=\frac{1}{n}\frac{\partial \lv \bm{e}_{-i} \rv^2 }{\partial \bm{x}_i} = \frac{2\bm{e}_{-i}}{n} \frac{\partial \bm{e}_{-i} }{\partial \bm{x}_i} .
\end{align*}

Taking derivatives of the constituent parts of $\bm{e}_{-i}$ in Equation \eqref{eq_press_refac} with respect to $\bm{x}_i$ we obtain
\begin{align}
\dx \bm{e}_i = -2\bm{\beta}\tr, 
~~
\dx t_i^2\bm{E}_{y,i}= 2\bm{E}_{y,i}\tr t_i \bm{u}\tr ,
~~
\dx \bm{b}_i = - 2\bm{\beta}\tr,
~~
\dx (1-t_i^2)(1-s_i^2) = -2t_i\bm{u}\tr.
\label{eq_dx1}
\end{align}

%
%

We can now obtain $\frac{\partial \bm{e}_{-i}}{\partial \bm{x}_i}$ by combining Equations \eqref{eq_dx1} 
so that 
\begin{align*}
\dx \bm{e}_{-i} 
 = & 2\frac{\left( - \bm{E}_{y,i}\tr  t_i \bm{u}\tr \right) }{(1-t_i^2)(1-s_i^2)} +2\bm{e}_{t,-i}\tr, 
\end{align*}
where $\bm{e}_{t,-i} = \frac{\bm{e}_{-i}}{(1-h_{t,i})}$.
Finally,
\begin{align*}
\frac{\partial J}{\partial \bm{x}_i} & = \frac{2\bm{e}_{-i}}{n} \left( 2\frac{\left( - \bm{E}_{y,i}\tr  t_i \bm{u}\tr \right) }{(1-t_i^2)(1-s_i^2)} +2\bm{e}_{t,-i}\tr \right) .
\end{align*}


The derivation of predictive influence with respect to $\bm{y}_i$
follows the same argument and so the predictive influence, $\bm{\pi}_{\bm{x},\bm{y}} (\bm{x}_i, \bm{y}_i) = [\frac{\partial J}{\partial \bm{x}_i} \tr, \frac{\partial J}{\partial \bm{y}_i} \tr] \tr$.

\bibliographystyle{abbrv}
\bibliography{press}

\begin{thebibliography}{10}

\bibitem{Belsley1980}
D.~Belsley and E.~Kuh.
\newblock {\em {Regression diagnostics: Identifying influential data and
  sources of collinearity}}.
\newblock Wiley, New York, New York, USA, 1 edition, 2004.

\bibitem{Bickel2004b}
P.~Bickel and E.~Levina.
\newblock Some theory for fisher's linear discriminant function, 'naive bayes',
  and some alternatives when there are many more variables than observations.
\newblock {\em Bernoulli}, 10(6):989--1010, 2004.

\bibitem{Bickel2004}
S.~Bickel and T.~Scheffer.
\newblock {Multi-view clustering}.
\newblock In {\em Proceedings of the IEEE international conference on data
  mining}. Citeseer, 2004.

\bibitem{Bickel2005}
S.~Bickel and T.~Scheffer.
\newblock {Estimation of mixture models using Co-EM}.
\newblock In {\em Machine Learning: ECML 2005}, volume 3720, pages 35--46.
  Springer, 2005.

\bibitem{Bishop2006}
C.~Bishop.
\newblock {\em {Pattern recognition and machine learning}}, volume~4.
\newblock Springer New York, Aug. 2006.

\bibitem{Bruno2009}
E.~Bruno and S.~Marchand-Maillet.
\newblock {Multiview clustering: A late fusion approach using latent models}.
\newblock In {\em Proceedings of the 32nd international ACM SIGIR conference on
  Research and development in information retrieval}, pages 736--737. ACM,
  2009.

\bibitem{Caruana1997}
R.~Caruana.
\newblock {Multitask learning}.
\newblock {\em Machine Learning}, 28(1):41--75, 1997.

\bibitem{Cesa-Bianchi2010}
N.~Cesa-Bianchi, D.~Hardoon, and G.~Leen.
\newblock {Guest Editorial: Learning from multiple sources}.
\newblock {\em Machine Learning}, 79(1):1--3, 2010.

\bibitem{Chaudhuri2009}
K.~Chaudhuri, S.~M. Kakade, K.~Livescu, and K.~Sridharan.
\newblock {Multi-view clustering via canonical correlation analysis}.
\newblock {\em Proceedings of the 26th Annual International Conference on
  Machine Learning - ICML '09}, pages 1--8, 2009.

\bibitem{Sa2010}
V.~de~Sa, P.~Gallagher, J.~Lewis, and V.~Malave.
\newblock {Multi-view kernel construction}.
\newblock {\em Machine learning}, 79(1):47--71, 2010.

\bibitem{Dudoit2000}
S.~Dudoit, J.~Fridlyand, and T.~Speed.
\newblock Comparison of discrimination methods for the classification of tumors
  using gene expression data.
\newblock Technical report, UC Berkeley, 2000.

\bibitem{Golub1996}
G.~Golub and C.~{Van Loan}.
\newblock {\em {Matrix computations}}.
\newblock Johns Hopkins Univ Pr, 1996.

\bibitem{Hou2010}
C.~Hou, C.~Zhang, Y.~Wu, and F.~Nie.
\newblock {Multiple view semi-supervised dimensionality reduction}.
\newblock {\em Pattern recognition}, 43(3):720--730, 2010.

\bibitem{Kannan2008}
R.~Kannan and S.~Vempala.
\newblock {Spectral Algorithms}.
\newblock {\em Foundations and Trends in Theoretical Computer Science},
  4(3--4):157--288, 2008.

\bibitem{Lange2006}
T.~Lange and J.~Buhmann.
\newblock {Fusion of similarity data in clustering}.
\newblock In {\em Advances in Neural Information Processing Systems},
  volume~18, page 723. Citeseer, 2006.

\bibitem{LeCao2009}
K.-A. {L\^{e} Cao}, P.~G.~P. Martin, C.~Robert-Grani\'{e}, and P.~Besse.
\newblock {Sparse canonical methods for biological data integration:
  application to a cross-platform study.}
\newblock {\em BMC bioinformatics}, 10:34, 2009.

\bibitem{Li}
T.~Li and C.~Ding.
\newblock {Weighted consensus clustering}.
\newblock In {\em Proceedings of the 2008 SIAM International Conference on Data
  Mining}, volume~1, pages 798--809. Citeseer, 2008.

\bibitem{Long2008}
B.~Long, P.~Yu, and Z.~Zhang.
\newblock {A general model for multiple view unsupervised learning}.
\newblock In {\em Proceedings of the 2008 SIAM International Conference on Data
  Mining}, 2008.

\bibitem{Martens2000}
H.~Martens.
\newblock {Modified Jack-knife estimation of parameter uncertainty in bilinear
  modelling by partial least squares regression (PLSR)}.
\newblock {\em Food Quality and Preference}, 11(1-2):5--16, Jan. 2000.

\bibitem{McWilliams2010a}
B.~McWilliams and G.~Montana.
\newblock {A PRESS statistic for two-block partial least squares regression}.
\newblock In {\em Computational Intelligence (UKCI), 2010 UK Workshop on},
  pages 1--6. IEEE, 2010.

\bibitem{McWilliams2010}
B.~McWilliams and G.~Montana.
\newblock {Sparse partial least squares regression for on-line variable
  selection with multivariate data streams}.
\newblock {\em Statistical Analysis and Data Mining}, 3(3):170--193, 2010.

\bibitem{Meloun2001}
M.~Meloun.
\newblock {Detection of single influential points in OLS regression model
  building}.
\newblock {\em Analytica Chimica Acta}, 439(2):169--191, July 2001.

\bibitem{Parkhomenko2009}
E.~Parkhomenko, D.~Tritchler, and J.~Beyene.
\newblock {Sparse Canonical Correlation Analysis with Application to Genomic
  Data Integration}.
\newblock {\em Statistical Applications in Genetics and Molecular Biology},
  8(1), 2009.

\bibitem{Pregibon1981}
D.~Pregibon.
\newblock {Logistic regression diagnostics}.
\newblock {\em The Annals of Statistics}, 9(4):705--724, 1981.

\bibitem{Rosipal2006}
R.~Rosipal and N.~Kr\"{a}mer.
\newblock {Overview and Recent Advances in Partial Least Squares}.
\newblock In {\em Subspace, Latent Structure and Feature Selection}, pages
  34--51. Springer, 2006.

\bibitem{Stone1974a}
M.~Stone.
\newblock {Cross-validation and multinomial prediction}.
\newblock {\em Biometrika}, 61(3):509--515, 1974.

\bibitem{Tian2006}
Y.~Tian, T.~Huang, and W.~Gao.
\newblock {Robust Collective Classification with Contextual Dependency Network
  Models}.
\newblock {\em Advanced Data Mining and Applications}, pages 173--180, 2006.

\bibitem{Tibshirani2003}
R.~Tibshirani, T.~Hastie, B.~Narasimhan, and G.~Chu.
\newblock Class prediction by nearest shrunken centroids, with applications to
  dna microarrays.
\newblock {\em Statistical Science}, 18(1):104--117, 2003.

\bibitem{Tzortzis2009}
G.~Tzortzis and A.~Likas.
\newblock {Convex mixture models for multi-view clustering}.
\newblock {\em Artificial Neural Networks--ICANN 2009}, pages 205--214, 2009.

\bibitem{Wegelin2000}
J.~Wegelin.
\newblock {A Survey of Partial Least Squares (PLS) Methods, with Emphasis on
  the Two-Block Case}.
\newblock Technical report, University of Washington, 2000.

\bibitem{Witten2009a}
D.~M. Witten, R.~Tibshirani, and T.~Hastie.
\newblock {A penalized matrix decomposition, with applications to sparse
  principal components and canonical correlation analysis.}
\newblock {\em Biostatistics (Oxford, England)}, 10(3):515--34, July 2009.

\bibitem{Wold1984}
S.~Wold, A.~Ruhe, H.~Wold, and W.~J. Dunn.
\newblock {The Collinearity Problem in Linear Regression. The Partial Least
  Squares (PLS) Approach to Generalized Inverses}.
\newblock {\em SIAM journal on Scientific Computing}, 5(3):735--743, 1984.

\end{thebibliography}

\newpage


\begin{table}[h]
\begin{center}
\begin{tabular}{|l|r|r|r|r|r|r|}
\hline
University & \multicolumn{4}{|c|}{Observations}      & View 1 & View 2 \\ \hline
           & Course & Student & Staff & Faculty &   (p)     &  (q)       \\ \hline
Cornell    & 83     & 18      & 38    &	32      & 1703   & 694    \\ \hline
Texas      & 103    & 18      & 33    &	31      & 1703   & 660    \\ \hline
Washington & 106    & 19      & 65    &	27      & 1703   & 715    \\ \hline
Wisconsin  & 116    & 22      & 70    &	34      & 1703   & 745   \\ \hline
\end{tabular}
\caption{A summary of the number of observations and variables in the different configurations of the WebKB dataset. \label{tab_webkb}}
\end{center}
\end{table}


\begin{table}
\begin{center}
\begin{tabular}{lrrrrr}
\hline
University 								  & Global PLS & WCC & MV-CCA  & MV-Kernel & MVPP 
\\ \hline
Cornell   								 &        & &         &           
    					\\
$~~$Acc										&	   -     &0.50  & 0.56 & 0.65    	 &0.96 
\\ 
$~~$Error									&163.69    & 46.71 & 137.65 & 159.93     &37.35 
\\ \hline
Texas      								&        & &        &           
\\
$~~$Acc										&   -     &0.50  & 0.57  & 0.71    	 &0.95  
\\ 
$~~$Error									&177.50   & 40.90  & 132.01 & 173.79     &33.74 
\\ \hline
Washington &        &         &    &       
 						\\
$~~$Acc										&   -     &0.87  & 0.79  & 0.69    	 &0.97   
 \\ 
$~~$Error									&209.40   & 46.44  & 106.86 & 109.16     &31.53	
\\ \hline
Wisconsin  &        &         &     &      
 						\\		
$~~$Acc										&    -     &0.67  & 0.76  & 0.59    	 &0.98
     \\ 
$~~$Error									&234.16   &  68.86 & 171.58 & 244.85     &55.72 
				\\ \hline
\end{tabular}
\caption{The clustering accuracies (Acc) and mean squared leave-one-out prediction error on the WebKB-2 dataset. MVPP consistently accurately recovers the true clusters and therefore also obtains the best prediction accuracy. The large variance in prediction accuracy between the other methods demonstrates the importance of fitting the correct local models. \label{tab_webkb_2}}
\end{center}
\end{table}


\begin{table}
\begin{center}
\begin{tabular}{lrrrrr}
\hline
University 								  & Global PLS & WCC & MV-CCA  & MV-Kernel & MVPP  
  		 \\ \hline
Cornell   								 &        & &         &           
\\
$~~$Acc										&	   -     &0.70  & 0.69 & 0.44    	 &0.83 
\\ 
$~~$Error									&163.69    & 35.43 & 19.77 & 105.04     &17.89        
 \\ \hline
Texas      								&        & &        &           
\\
$~~$Acc										&    -    &0.58  & 0.68  & 0.41    	 &0.86  
\\ 
$~~$Error									&177.50   & 54.35  & 26.21 & 141.34     &18.97 				
\\ \hline
Washington &        &         &    &       
\\
$~~$Acc										&    -    &0.70  & 0.68  & 0.53    	 &0.75   
 \\ 
$~~$Error									&209.40   & 36.14  & 33.26 & 98.58     &17.34				
\\ \hline
Wisconsin  &        &         &     &      
	\\		
$~~$Acc										&   -      &0.69  & 0.74  & 0.53    	 &0.85 
 \\ 
$~~$Error									&234.16   &  61.61 & 31.27 & 110.89     &21.13 				
\\ \hline
\end{tabular}
\caption{The clustering accuracies (Acc) and mean squared leave-one-out prediction error on the WebKB-4 dataset. MVPP again achieves the best clustering and prediction performance. Although the clustering accuracy is worse than in the WebKB-2 configuration, the improved prediction performance suggests that fitting four clusters is a more accurate model of the data. \label{tab_webkb_4}}
\end{center}
\end{table}


\begin{table}
\begin{center}
\begin{tabular}{lrrrrr}
\hline
Configuration 								  & Global PLS& WCC & MV-CCA  & MV-Kernel & MVPP   		 
\\ \hline
Text + Inbound   								 &        & &        &           &    
\\
$~~$Acc													 &  - & 0.76   & 0.76 & 0.73    	 &0.81  
\\ 
$~~$Error													&344.06    & 70.62 & 76.50 &110.30     &39.51        
\\ \hline
Text + Outbound      								&        & &        &           & 
\\
$~~$Acc														&  -& 0.76    & 0.76 & 0.72    	 &0.87  
\\ 
$~~$Error													&278.53   & 110.46  & 84.50 & 73.95     &52.96 			
\\ \hline
\end{tabular}
\caption{The clustering accuracies (Acc) and mean squared leave-one-out prediction error on the Citeseer dataset. MVPP achieves the best clustering accuracy and prediction error whereas the other methods all achieve a similar clustering accuracy.  \label{tab_citeseer_6}}
\end{center}
\end{table}

\end{document}